\documentclass[10pt,twocolumn,letterpaper]{article}

\usepackage{iccv}              
\PassOptionsToPackage{hyphens}{url}
\usepackage[pagebackref,breaklinks,colorlinks,allcolors=iccvblue]{hyperref}
\usepackage{tabularray}
\usepackage{booktabs} 
\usepackage{graphicx}
\usepackage{multirow}
\usepackage{multicol}
\usepackage{makecell}
\usepackage{caption}
\usepackage{adjustbox}
\usepackage{float}
\usepackage{enumitem}
\usepackage{wrapfig}
\usepackage[table]{xcolor}
\usepackage{amssymb}
\usepackage{float}

\usepackage{pifont}
\newcommand{\cmark}{\ding{51}}%
\newcommand{\xmark}{\ding{55}}%

\newcommand{\vf}{\mathbf{f}}

\usepackage{algorithm}
\usepackage{algpseudocode}
\usepackage{amsmath, amssymb}
\usepackage[accsupp]{axessibility}  

\definecolor{iccvblue}{rgb}{0.21,0.49,0.74}
\usepackage[pagebackref,breaklinks,colorlinks,allcolors=iccvblue]{hyperref}

\def\paperID{2360} 
\def\confName{ICCV}
\def\confYear{2025}

\title{MaskControl: Spatio-Temporal Control for Masked Motion Synthesis}

\author{Ekkasit Pinyoanuntapong$^1$, Muhammad Usama Saleem$^1$, Korrawe Karunratanakul$^2$, \\ Pu Wang$^1$, Hongfei Xue$^1$, Chen Chen$^3$, Chuan Guo$^4$, Junli Cao$^4$, Jian Ren$^4$, Sergey Tulyakov$^4$  \\
{\bfseries\fontsize{9pt}{10pt}\selectfont
$^1$University of North Carolina at Charlotte \quad 
$^2$ETH Zürich \quad 
$^3$University of Central Florida \quad 
$^4$Snap Inc.}\\
\texttt{\small epinyoan@charlotte.edu} \\
}

\begin{document}
\let\oldtwocolumn\twocolumn
\renewcommand\twocolumn[1][]{%
    \oldtwocolumn[{#1}{
    \begin{center}
           \includegraphics[width=\textwidth]{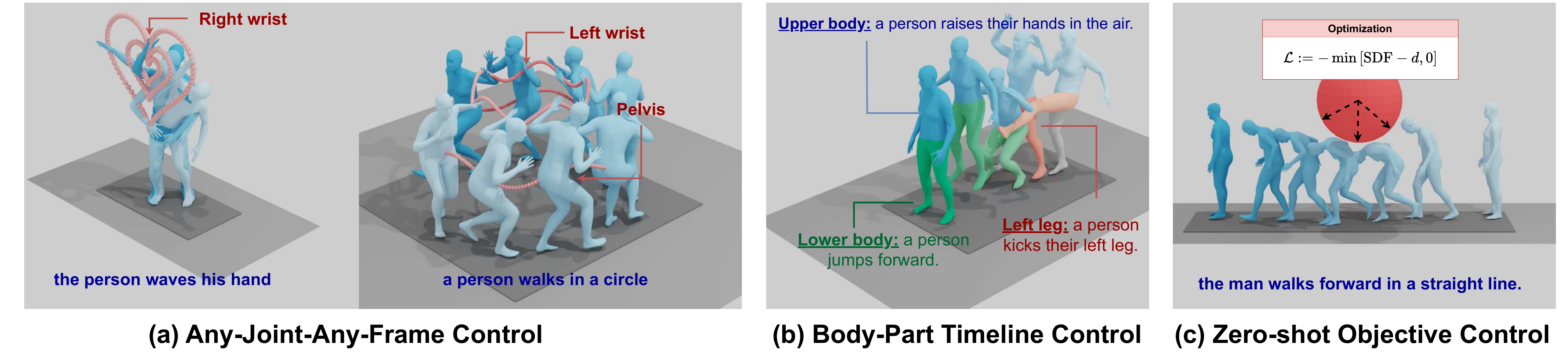}
           \vspace{-10pt}
           \captionof{figure}{MaskControl enables a wide range of applications in text-to-motion generation with high quality and precision. (a) Any-Joint Any-Frame Control: Controls specific joints at designated frames. (b) Body-Part Timeline Control: Generates motion from multiple text prompts, each corresponding to different body parts. (c) Arbitrary Objective Control: Supports any control function at inference time.}
           \label{fig:landing}
        \end{center}
        
    }]
}

\maketitle
\begin{abstract}
Recent advances in motion diffusion models have enabled spatially controllable text-to-motion generation. However, these models struggle to achieve high-precision control while maintaining high-quality motion generation. To address these challenges, we propose MaskControl, the first approach to introduce controllability to the generative masked motion model. Our approach introduces two key innovations. First, \textit{Logits Regularizer} implicitly perturbs logits at training time to align the distribution of motion tokens with the controlled joint positions, while regularizing the categorical token prediction to ensure high-fidelity generation. Second, \textit{Logit Optimization} explicitly optimizes the predicted logits during inference time, directly reshaping the token distribution that forces the generated motion to accurately align with the controlled joint positions. Moreover,  we introduce \textit{Differentiable Expectation Sampling (DES)} to combat the non-differential distribution sampling process encountered by logits regularizer and optimization. Extensive experiments demonstrate that MaskControl outperforms state-of-the-art methods, achieving superior motion quality (FID decreases by ~77\%) and higher control precision (average error 0.91 vs. 1.08). Additionally, MaskControl enables diverse applications, including any-joint-any-frame control, body-part timeline control, and zero-shot objective control. Video visualization can be found at \url{https://www.ekkasit.com/ControlMM-page/}

\end{abstract}

\vspace{-10pt}
\section{Introduction}

Text-driven human motion generation has recently gained significant attention due to the semantic richness and intuitive nature of natural language descriptions. This approach has broad applications in animation, film, virtual/augmented reality (VR/AR), and robotics. While text descriptions offer a wealth of semantic guidance for motion generation, they often fall short in providing precise joint control over specific human joints, such as the pelvis and hands. As a result, achieving natural interaction with the environment and fluid navigation through 3D space remains a challenge.


To tackle this challenge, a few controllable motion generation models have been developed recently to synthesize realistic human movements that align with both text prompts and joint control signals \cite{PriorMDM, trace, OmniControl, TLcontrol}. However, existing solutions face significant difficulties in generating high-fidelity motion with precise and flexible joint control. In particular, current models struggle to support both sparse and dense joint control signals simultaneously. For instance,  some models excel at generating natural human movements that traverse sparse waypoints \cite{GMD, trace}, while others are more effective at synthesizing motions that follow detailed trajectories specifying human positions at each time point \cite{TLcontrol}. Recent attempts to support both sparse and dense joint inputs encounter issues with control precision; the generated motion often is not aligned well with the control conditions \cite{OmniControl}. Besides unsatisfied joint flexibility and accuracy,  the quality of motion generation in controllable models remains suboptimal, as evidenced by much worse FID scores compared to models that rely solely on text inputs. Moreover, most current methods utilize motion-space diffusion models, applying diffusion processes directly to raw motion sequences. While this design facilitates the incorporation of joint control signals, the redundancy in raw data introduces computational overhead, resulting in slower motion generation speeds.

Current controllable motion generation methods mostly rely on diffusion model \citep{OmniControl, motionlcm, GMD, OmniControl}. However, these methods exhibit lower control precision and less generation quality. The current SOTA method, TLControl,  \cite{TLcontrol} achieves high-precision control through test-time optimization with a simple feedforward transformer, but this comes at the cost of lower generation quality, as shown in Tab. \ref{tab:overall_sota}. Moreover, these methods cannot adapt to arbitrary objective functions at inference time in a zero-shot manner.


To address these challenges, we introduce the first method that integrates joint control into generative masked motion models, enabling simultaneous high-fidelity motion generation and precise control across a broad range of tasks, as illustrated in Fig. \ref{fig:landing}. In contrast to diffusion-based approaches, masked models generate motion sequences by training a multi-category token classifier and subsequently sampling from the learned categorical distributions conditioned on input signals, such as text and joint trajectories. Building upon this insight, we propose MaskControl—a straightforward yet powerful control mechanism for generative masked motion models—that implicitly and explicitly manipulates the logits of the token classifier to align token distributions closely with input control signals.

Our contributions can be summarized as follows. 
\begin{itemize}
    \item We propose MaskControl, the first approach to introduce controllability to generative masked motion models.
    \item We introduce two novel control components of MaskControl: (1) \textit{Logits Regularizer}  that implicitly perturbs logits at training time to align the distribution of motion tokens with the controlled joint positions. (2) \textit{Logits Optimization} explicitly optimizes the predicted logits during inference time, directly reshaping the token distribution to minimize the residual errors between the generated motion and the target joint position. Moreover, we propose \textit{Differentiable Expectation Sampling (DES)} to overcome the challenge of non-differentiable probabilistic token selection in logits regularization and optimization.
    \item We show that Logits Optimization can be generalized to solve unseen control tasks in a zero-shot manner. 
    \item We conduct extensive qualitative and quantitative evaluations on multiple tasks, demonstrating that our approach outperforms current SOTA in both motion generation quality and control precision while supporting multiple applications \textit{i.e.} any-joint-any-frame control, body part timeline control, and zero-shot objective control.
\end{itemize}

\begin{table}[h]
\centering
\caption{Comparison of text-conditioned motion generation with joint control signals. Our MaskControl SOTA achieves performance by leveraging Masked Motion Model, demonstrating high-precision control (low Average Error) while maintaining high generation quality (low FID). Previous SOTA methods utilize diffusion in motion space, latent space, or simple feed-forward models. `\textcolor{red}{\textbf{\cmark}}' indicates the ability to control motion using zero-shot objective functions, while `\xmark' denotes the lack of this capability. `-' signifies control limited to the pelvis.}
\label{tab:overall_sota}
\scalebox{0.72}{
\begin{tabular}{lcccccc}
\toprule
\textbf{Method} & \textbf{FID $\downarrow$} & \textbf{\makecell{Average\\Error (cm) $\downarrow$}} & \textbf{Base Model} & \textbf{\makecell{Zero-shot\\Objective}} \\
\midrule
GMD \cite{GMD} & 0.576 & 14.39 & Motion Diffusion & - \\
OmniControl \cite{OmniControl} & 0.218 & 3.38 & Motion Diffusion & \xmark \\
MotionLCM \cite{motionlcm} & 0.531 & 18.97 & Latent Diffusion & \xmark \\
TLControl \cite{TLcontrol} & 0.271 & 1.08 & Feed Forward & \xmark \\
\textbf{Ours} & \textcolor{red}{\textbf{0.061}} & \textcolor{red}{\textbf{0.98}} & \textbf{Masked Model} & \textcolor{red}{\textbf{\cmark}} \\
\bottomrule
\end{tabular}}
\vspace{-10pt}
\end{table}

\section{Related Work}

\textbf{Text-driven Motion Generation.}
Early methods for text-to-motion generation primarily focus on aligning the latent distributions of motion and language, typically by employing loss functions such as Kullback-Leibler (KL) divergence and contrastive losses. Representative works in this domain include Language2Pose \citep{Language2Pose}, TEMOS \citep{TEMOS}, T2M \citep{t2m}, MotionCLIP \citep{MotionCLIP}, and DropTriple \citep{CrossModalRF}. However, the inherent discrepancy between the distribution of text and motion often results in suboptimal generation quality when using these latent space alignment techniques.

Recently, diffusion models have become a widespread choice for text-to-motion generation, operating directly in the motion space~\citep{MDM,MotionDiffuse,FLAME}, VAE latent space~\citep{MLD}, or quantized space~\citep{DiverseMotion,kong2023priority}. In these works, the model gradually denoises the whole motion sequence to generate the output in the reverse diffusion process. Another line of work explores the token-based models in the human motion domain, for example, autoregressive GPTs~\citep{TM2T,T2M-GPT,MotionGPT,AttT2M} and \textit{masked motion modeling}~\citep{MMM, BAMM, momask}. These methods learn to generate discrete motion token sequences that are obtained from a pretrained motion VQVAE~\citep{vqgan,Hierarchical-vqvae}. While GPT models usually predict the next token from history tokens, masked motion models utilize the bidirectional context to decode the masked motion tokens. By predicting multiple tokens at once, the masked modeling methods can generate motion sequences in as few as 10 steps, achieving state-of-the-art performance on generation quality and efficiency. Despite the performance gains of masked motion models, supporting spatial controllability in these models remains unexploited. This paper is the first work that proposes controllable masked motion model to simultaneously achieve high-quality motion generation with high-precision spatial control.


\textbf{Controllable Motion Synthesis.}
In addition to text prompts, synthesizing motion based on other control signals has also been a topic of interest. Example control modalities include music \citep{AIChoreographer, DanceFormer, Dancingtomusic, Bailando, Bailando++, EDGE}, interacting object \citep{Nifty,CgHoi,CHOIS, Text2HOI}, tracking sensors \citep{AGRoL}, scene \citep{SceneDiffuser, MoveasYouSay} programmable motion \citep{programmable}, style \citep{SMooDi}, goal-reaching task \citep{wandr}, and multi-track timeline control \citep{Multi-TrackTimelineControl}. \cite{AMP, ASE, PhysicsbasedHM, PhysDiff, PHC, pulse, maskedmimic} incorporate physics into motion generation. To control the trajectory, PriorMDM \citep{PriorMDM} finetunes MDM to enable control over the locations of end effectors. CondMDI \citep{cohan2024flexible} generates motion in-betweening from arbitrarily placed dense or sparse keyframes. GMD \citep{GMD} and Trace and Pace \citep{trace} incorporates spatial control into the diffusion process by guiding the root joint location. OmniControl \citep{OmniControl} extends the control framework to any joint, while MotionLCM \citep{motionlcm} applies this control in the latent space, both leveraging ControlNet \citep{ControlNet}. DNO \citep{DNO} introduces an optimization process on the diffusion noise to generate motion that minimizes a differentiable objective function. Recent approaches \citep{TLcontrol, CoMo} model each body part separately to achieve fine-grained control but are limited to dense trajectory objectives. The existing frameworks for controllable motion generation predominantly rely on diffusion models; however, these models typically suffer from low-quality motion generation.

\begin{figure*}[ht]
 \centering
  \includegraphics[width=1\textwidth]{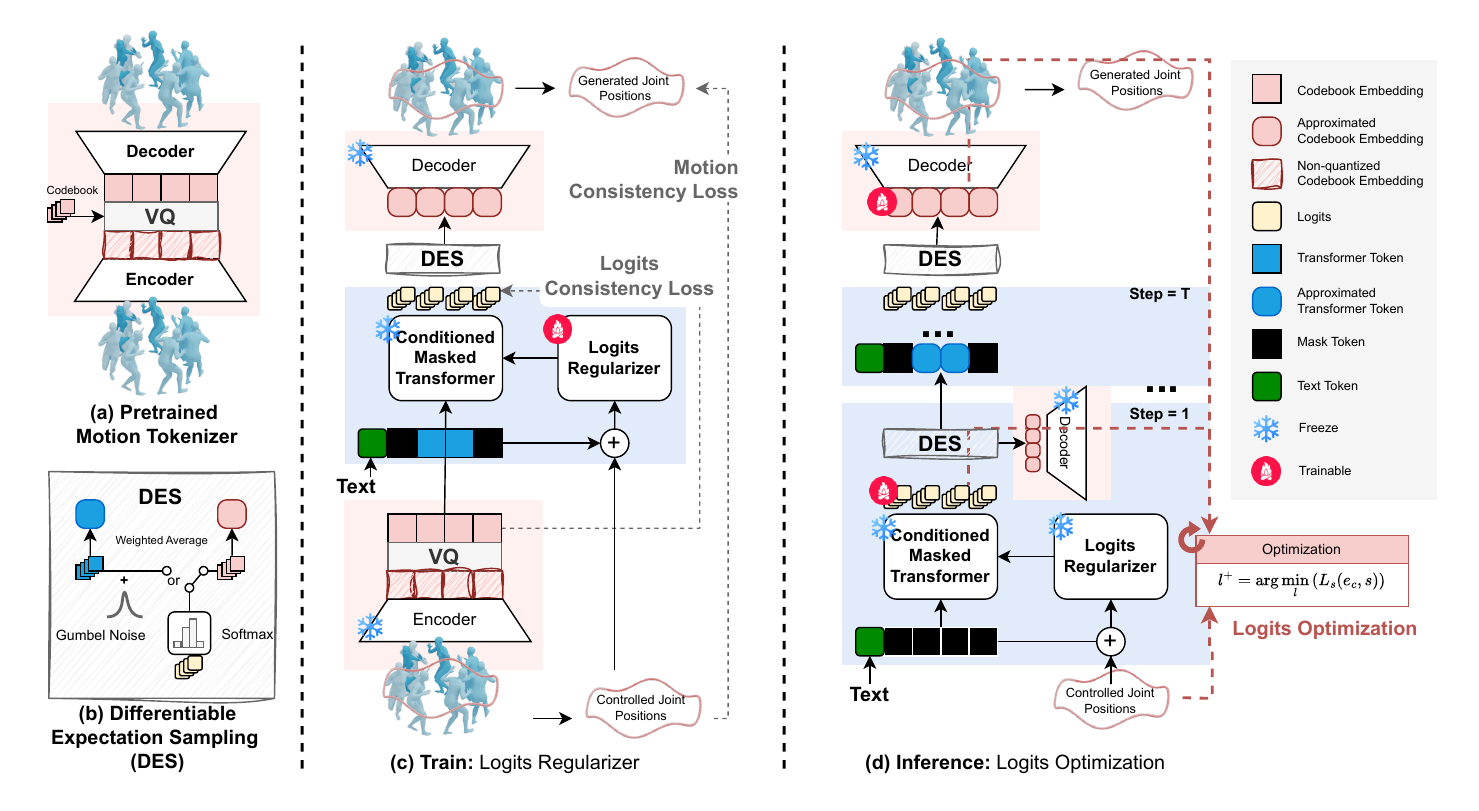}
  \vspace{-10pt}
  \caption{Overall architecture of MaskControl. \textbf{\textit{(a) Motion Tokenizer}} transforms the motion sequence into discrete motion tokens. \textbf{\textit{(b) Differentiable Expectation Sampling (DES)}} is a differentiable sampling from logits enabling differentiable conversion between discrete tokens in codebook space and transformer token space. \textbf{\textit{(c) Training: Logits Regularizer}} ensures high-quality motion by generating embedding closely aligns with joint control signals during an unmasking process. \textbf{\textit{(d) Inference: Logits Optimization}} guides logits during the unmasking process at inference time based on the objective function.
  }
  \label{fig:architecture}
\vspace{-10pt}
\end{figure*}

\section{MaskControl}
\label{sec:method}
The objective of MaskControl is to enable controllable text-to-motion generation based on a masked motion model that generates high-precision and high-quality motion. In particular, given a text prompt and an additional joint control signal, our goal is to generate a physically plausible human motion sequence that closely aligns with the textual descriptions, while following the joint control conditions, i.e., $(x,y,z)$ positions of each human joint at each frame in the motion sequence. Towards this goal, in Sec. \ref{sec:preliminary}, we first introduce the background of conditional motion synthesis based on the generative masked motion model. We then describe two key components of MaskControl, including \textit{Logits Regularizer} in Sec. \ref{sec:mask_consistency} and inference-time \textit{Logits Optimization} in Sec. \ref{sec:logit_edit}. The first component aims to learn the categorical distribution of motion tokens, conditioned on joint control signals during training time. The second component aims to improve control precision by optimally modifying learned motion distribution via \textit{Logits Optimization} during inference time. Lastly, we introduce \textit{Differentiable Expectation Sampling} to overcome non-differentiability of categorical sampling during Logits Regularizer and Optimization. 

\subsection{Preliminary: Masked Motion Model}
\label{sec:preliminary}
Generative masked motion models ~\citep{MMM, BAMM, momask} generally consist of two stages: \textit{Motion Tokenizer} and \textit{Text-conditioned Masked Transformer}. The objective of the Motion Tokenizer is to learn a discrete representation of motion by quantizing the encoder's output embedding $z$ into a codebook $\mathcal{C}$. For a given motion sequence $\mathcal{P} = [p_1, p_2, ..., p_F ]$, where each frame $p$ represents a 3D pose, \textit{Motion Tokenizer outputs} a discrete motion tokens $X = [x_1, x_2, ..., x_L ]$. Specifically, the encoder compresses $\mathcal{P}$ into a latent embedding $z \in \mathbb{R}^{t \times d}$ with a downsampling rate of $F/L$. The embedding $z$ is quantized into codes $c \in \mathcal{C}$ from the codebook $\mathcal{C} = \{c_k\}_{k=1}^K$, which contains $K$ codes. The nearest code is selected by minimizing the Euclidean distance between $z$ and the codebook entries, computed as $\hat{z} = \operatorname{argmin}_j \left\| \mathbf{z} - c_j \right\|_2^2$. The vector quantization loss $L_{VQ}$ is defined as:
\begin{equation}
L_{VQ} = \|\operatorname{sg}(\mathbf{z}) - \mathbf{c_j}\|_2^2 + \beta \|\mathbf{z} - \operatorname{sg}(\mathbf{c_j})\|_2^2,
\end{equation}
where $\operatorname{sg}(\cdot)$ is the stop-gradient operator and $\beta$ is a hyper-parameter for commitment loss.

During the second stage,  the quantized motion token sequence  $X = [x_1, x_2, ..., x_L ]$ is updated with \texttt{[MASK]} tokens to form the corrupted motion sequence $X_{\overline{\mathbf{M}}}$. This corrupted sequence along with text embedding $W$ are fed into a text-conditioned masked transformer parameterized by $\theta$ to reconstruct input motion token sequence with reconstruction probability equal to $p_{\theta}\left(x_i \mid X_{\overline{\mathbf{M}}}, W\right)$, which is obtained by the motion token classifier. The objective is to minimize the negative log-likelihood of the predicted masked tokens conditioned on text: 
\begin{equation}
\label{eq:mmm_eq}
\small
\mathcal{L}_{\text {mask}}=-\underset{\mathbf{X} \in \mathcal{D}}{\mathbb{E}}\left[\sum_{\forall i \in[1, L]} \log p\left(x_i \mid X_{\overline{\mathbf{M}}}, W\right)\right].
\end{equation}

During inference, the transformer masks out the tokens with the least confidence and re-samples these tokens according to their respective distributions $p_{\theta}\left(x_i \mid X_{\overline{\mathbf{M}}}, W\right)$in the subsequent step. The number of masked tokens $n_M$ is controlled by a masking schedule, a decaying function of the step $t$. Early steps use a large masking ratio due to high uncertainty, and as the process continues, the ratio decreases as more context is available from previous predictions.


\subsection{Logits Regularizer} 
\label{sec:mask_consistency}
MaskControl aims to generate a human motion sequence based on the text prompt $W$ and joint control signals $S$.  Towards this goal, we introduce \textit{Logits Regularizer} to conditioned masked transformer, which aims to learn the motion token distribution jointly conditioned both on $W$ and $S$. \textit{Logits Regularizer} implicitly alters the output logits of pre-trained text-to-motion model, thus changing the distribution of motion tokens, toward joint control positions.


\setlength{\parindent}{0pt}
\textbf{Model Architecture.}   
Diffusion ControlNet has shown its excellence in adding additional control signals to pretrained image diffusion model \citep{ControlNet}. We demonstrate for the first time that similar design principle can be applied to introduce controllability to generative masked models \citep{BERT, MaskGIT, Muse, Phenaki}.  
Our architecture consists of a pre-trained text-conditioned masked motion model and a \textit{Logits Regularizer}. The pre-trained model provides a strong motion prior based on text prompts, while the \textit{Logits Regularizer} introduces additional joint control signals. Specifically, the \textit{Logits Regularizer} is a trainable replica of the pre-trained masked motion model, as shown in Fig \ref{fig:architecture}. Each Transformer layer in the original model is paired with a corresponding layer in the trainable copy, connected via a zero-initialized linear layer. This initialization ensures that the layers have no effect at the start of training. Unlike the original masked motion model, the \textit{Logits Regularizer} incorporates two conditions: the text prompt $W$ from the pre-trained CLIP model \citep{CLIP} and the joint control signals $S$. The text prompt $W$ influences the motion tokens through attention, while the joint control signal $S$ is directly added to the motion token sequence via a projection layer. 

\setlength{\parindent}{0pt}
\textbf{Joint Control Signal.}  The conditioned masked transformer is trained to learn the conditional distribution $p_{\theta}\left(x_k \mid X, X_{\overline{\mathbf{M}}}, W, S\right)$ by reconstructing the masked motion tokens $X_{\overline{\mathbf{M}}}$, conditioned on the unmasked tokens $X$, text prompt W, and joint control signals S. The joint control condition is a sequence of joint control signals $S = [s_1, s_2,...,s_F ]$ with $s_i \in \mathbb{R}^{j\times3}$. Each control signal $s_i$ specifies the targeted 3D coordinates of the joints to be controlled, among the total $j$ joints, while joints that are not controlled are zeroed out. Since the semantics of the generated motion are primarily influenced by the textual description, to guarantee the controllability of joint control signals, we extract the joint control signals from the generated motion sequence and directly optimize the consistency loss between input control signals and those extracted from the output. Note that zero-valued in 3D joint coordinates can be ambiguous. Please refer to Sec. \ref{sec:challenges_motioncontrol} for more details.

\setlength{\parindent}{0pt}
\textbf{Motion Consistency Loss} evaluates the alignment between the generated motion and the input joint control signals \( s \):
\begin{equation}
L_s(e_c, s) = \frac{\sum_n \sum_j \sigma_{nj} \odot \left\lVert s_{nj} - R(D(e_c)) \right\rVert}{\sum_n \sum_j \sigma_{nj}},
\label{dis}
\end{equation}
where \( \sigma_{nj} \) is a binary value indicating whether the joint control signals \( s \) contains a control value at frame \( n \) for joint \( j \). The motion tokenizer decoder \( D(\cdot) \) converts motion embedding into relative position in local coordinate system and \( R(\cdot) \) further transforms the joint's local positions to global absolute locations. 
The global location of the pelvis at a specific frame can be calculated from the cumulative aggregation of rotations and translations from all previous frames. The locations of the other joints can also be computed by the aggregation of the relative positions of the other joints to the pelvis position. 

\setlength{\parindent}{0pt}
\textbf{Logits Consistency Loss} extends the objective function of masked motion model in Eq. \ref{eq:mmm_eq} by applying negative log-likelihood to all positions, including unmasked ones, conditioned on both the text $W$ and joint control signals $S$.

\begin{equation}
\label{eq:logits}
\small
\mathcal{L}_{\text {logits}}=-\sum_{\forall i \in[1, L]} \log p\left(x_i \mid X_{\overline{\mathbf{M}}}, W, S\right).
\end{equation}

The final loss is the weighted combination \textit{Logits Consistency Loss} and \textit{Motion Consistency Loss}: 
\begin{equation}
\label{eq:all_controll_loss}
\mathcal{L} = \alpha \mathcal{L}_{\text{logits}} + (1-\alpha) L_s(e_c, s).
\end{equation}


\subsection{Logits Optimization} 
\label{sec:logit_edit}

The goal of inference-time \textit{Logits Optimization} is to enhance control precision by further reducing the discrepancy between the generated motion and the desired control objectives. This approach does not require pretraining on specific tasks, allowing the model to handle arbitrary objective functions during inference, enabling new control tasks in a zero-shot manner.

The core idea behind \textit{Logits Optimization} is to update the learned logits through gradient-guided optimization during inference, perturbing the motion token distribution. Optimizing logits makes optimization during unmasking process possible since the process requires conversion between transformer token for unmask sampling and codebook embedding for motion reconstruction, as shown in Fig. \ref{fig:architecture}. The optimization process is initialized with the logits obtained from conditioned masked transformer with \textit{Logits Regularizer}, and these logits are iteratively updated to minimize the motion consistency loss $L_s$.
\begin{equation}
l^+ = \arg\min_{l} \left( L_s(e_c, s) \right).
\label{eq:optimize_query}
\end{equation}
At each iteration \( m \), the logits \(l_m\) are updated using the following gradient-based approach:
    \begin{equation}
    \label{eq:TTT_logits}
    l_{m+1} = l_m - \eta \nabla_{l_m} L_s(l_m, s),
    \end{equation}
where \( \eta \) controls the magnitude of the updates to the logits, while \(  L_s(l_m, s) \) represents the gradient of the objective function w.r.t. the logits \(l_m\). This refinement process continues over $I$ iterations. Similarly, in the last unmask step, optimizing embeddings from the codebook space directly is possible. We can directly optimize the embedding: 
    \begin{equation}
    \label{eq:TTT_emb}
    e_{m+1} = e_m - \eta \nabla_{e_m} L_s(e_m, s),
    \end{equation}
where $e$ represents the embedding in the codebook space. For zero-shot objective control, loss function $L_s$ can be replaced with any arbitrary objective function.

\subsection{Differentiable Expectation Sampling (DES)} 
\label{sec:des}

\setlength{\parindent}{0pt}
\textbf{Differentiable Sampling.} 
Both logits regularizer and optimization update the logits by computing the gradient of the disparity between spatial-temporal control signals and generated motion sequence defined in eq. \ref{dis}. This requires sampling motion tokens according to the categorical token distribution during training/optimization time, which is inherently non-differentiable. To address this, we apply \textit{Straight-Through Gumbel-Softmax Estimator} \citep{gumbelsoftmax}, a reparameterization trick that allows differentiable sampling from a categorical distribution, i.e.,
\begin{equation}
\label{eq:gumbel}
p_{\theta}\left(x_k \mid X_{\overline{\mathbf{M}}}, W, S\right) = \frac{\exp\left( (\ell_k + g_k)/\tau\right)}{\sum_{j=1}^{K} \exp\left(\ell_j + g_j/\tau\right)},
\end{equation}
where $l$ is logits, $\tau$ refers to temperature, and $g$ represents Gumbel noise with $g_1, \dots, g_K$ being independent and identically distributed (i.i.d.) samples from a $\text{Gumbel}(0, 1)$ distribution. The $\text{Gumbel}(0, 1)$ distribution can be sampled via inverse transform sampling by first drawing $u \sim \text{Uniform}(0, 1)$ and then computing $g = -\log(-\log(u))$.

\setlength{\parindent}{0pt}
\textbf{Token Expectation.} Masked models reconstruct an embedding from the reconstruction probability $p_{\theta}(x_k \mid \cdot)$ by querying the discrete token embedding using the index with the highest probability, $\operatorname{argmax}_k p_{\theta}(x_k \mid \cdot)$. However, $\operatorname{argmax}$ is non-differentiable, and cannot be controlled by perturbation from a control signal. To address this, we relax the quantization constraint by utilizing the weighted average of the codebooks (or transformer tokens) w.r.t. the reconstruction probability $p_{\theta}(x_k \mid \cdot)$ to approximate continuous tokens. This can be expressed as the expectation of the reconstructed embedding $X_{recon}$:

\begin{equation}
\label{eq:expectation_embedding}
\mathbb{E}[X_{recon}] = \sum_{k=1}^{K} p_{\theta}\left(x_k \mid X_{\overline{\mathbf{M}}}, W, S\right) \cdot c_k.
\end{equation}

\begin{figure*}[ht]
 \centering
  \includegraphics[width=1\textwidth]{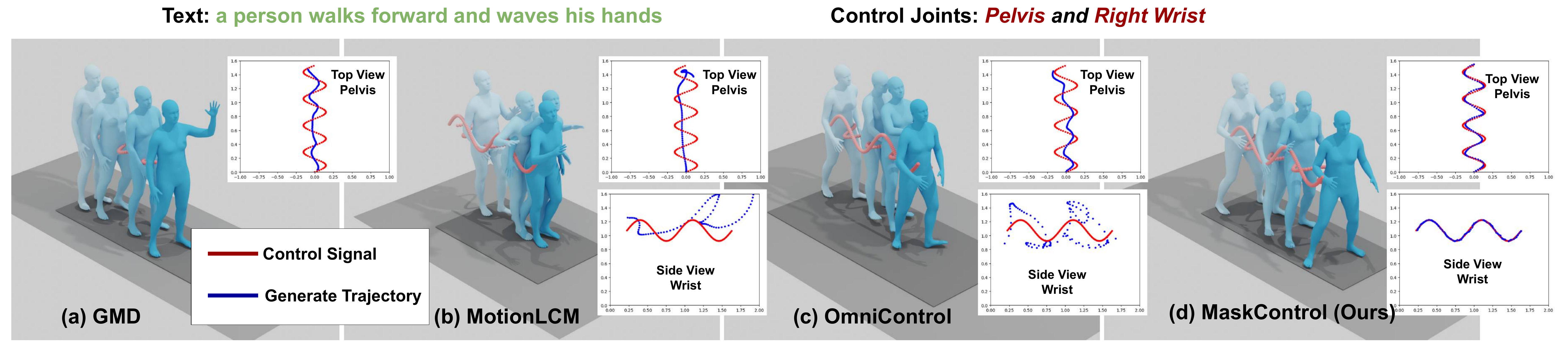}
  \caption{Visualization comparisons to state-of-the-art methods for any-joint any-frame control. The plots on the top display the top view of pelvis control (root trajectory), while the bottom plot shows the side view of the right wrist. \textcolor{red}{Red} represents the control signal, and \textcolor{blue}{Blue} represents the generated joint motion. }
  \label{fig:compare_sota}
  \vspace{-15pt}
\end{figure*}

\begin{figure}[ht]
 \centering
  \includegraphics[width=.5\textwidth]{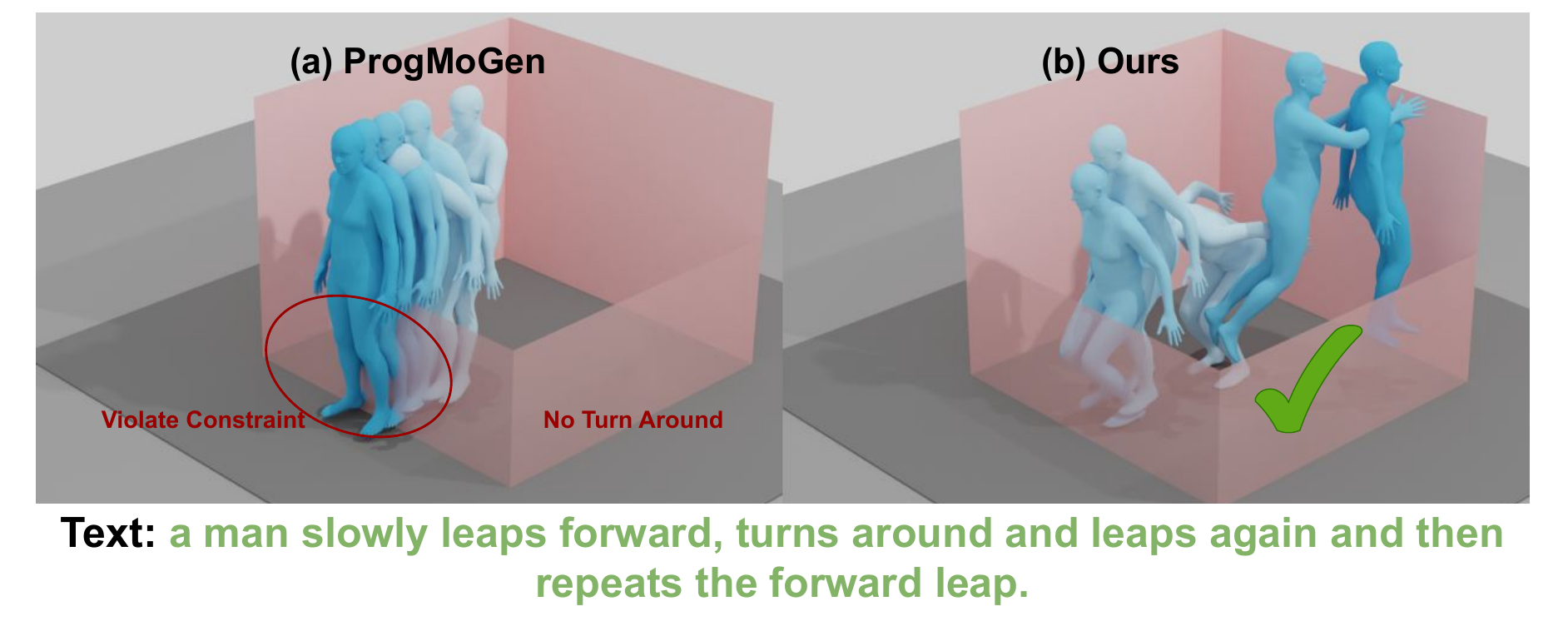}
  \vspace{-20pt}
  \caption{Visualization comparisons to state-of-the-art methods for zero-shot objective control. Objective: constrain a human to walk inside a square area. }
  \label{fig:zeroshot-sota}
  \vspace{-10pt}
\end{figure}

\section{Applications}
\textbf{Any-Joints-Any-Frame Control.}
To control specific joints at particular frames, the joint control signal can be directly applied to the desired joint and frame in the global position, as the loss function during training is specifically designed for this task.



\setlength{\parindent}{0pt}
\textbf{Body-Part Timeline Control.}
MaskControl supports motion generation conditioned on multiple joints, enabling control over body parts. To handle multiple prompts corresponding to various body parts and timelines, MaskControl processes each prompt sequentially. Initially, it generates motion without any body part control, then iteratively refines the motion by incorporating prompts conditioned on the specified body parts and timeline constraints from the prior generation. Since MaskControl allows joint control signals to target any joint and frame, partial body or temporal frame control is applicable within this framework. The detail of this process is described in \ref{sec:bodypart}.

\setlength{\parindent}{0pt}
\textbf{Zero-shot Objective Control.} 
In many human motion generation tasks, motion generation requires adaptation to dynamic constraints at inference time, such as human-scene interaction, human-object interaction, or human self-contact. \textit{Logits Optimization} allows optimization with arbitrary loss functions that take joints and frames as inputs, enabling zero-shot adaptation at inference time. However, direct optimization to satisfy arbitrary objective functions can often lead to unrealistic motion. To mitigate this, \textit{Logits Optimization} perturbs the logits during the unmasking process, allowing the masked transformer to re-predict them. By doing so, it ensures that the generated motion remains close to the learned distribution while still achieving the desired control.


\section{Experiment}

\begin{table*}[!ht]
\centering
\caption{Comparison of text-condition motion generation with joint control signal on the HumanML3D. The first section, ``Train on Pelvis Only," evaluates our model that was trained solely on the pelvis. The last section, ``Train on All Joints", is trained on all joints and reports the average evaluation for each joint. $\rightarrow$ indicates the closer to the real value, the better.}
\label{tab:main_eval}
\scalebox{0.74}{
\begin{tabular}{lcclccccc}
\toprule
\textbf{Method}                  & \textbf{Joint}       & \textbf{\makecell{R-Precision \\ Top-3 $\uparrow$}} & \textbf{FID $\downarrow$}   & \textbf{Diversity $\rightarrow$} & \textbf{\makecell{Foot Skating \\ Ratio $\downarrow$}} & \textbf{\makecell{Trajectory Error \\(\textgreater 50 cm) (\% ) $\downarrow$}} & \textbf{\makecell{Location Error \\(\textgreater 50 cm) (\% ) $\downarrow$}} & \textbf{\makecell{Average Error \\ (cm) $\downarrow$}}\\
\midrule
Real & - & 0.797 & 0.002 & 9.503 & - & 0.00 & 0.00 & 0.00 \\
\midrule

\noalign{\vspace{-6pt}}
\multicolumn{9}{c}{\cellcolor{gray!10}\textbf{Train on Pelvis Only}} \\
MDM &  & 0.602 & 0.698 & 9.197 & 0.1019 & 40.22 & 30.76 & 59.59 \\
PriorMDM &  & 0.583 & 0.475 & 9.156 & 0.0897 & 34.57 & 21.32 & 44.17 \\
GMD &  & 0.665 & 0.576 & 9.206 & 0.1009 & 9.31 & 3.21 & 14.39 \\
OmniControl (on pelvis)&  Pelvis & 0.687 & 0.218 & 9.422 & 0.0547 & 3.87 & 0.96 & 3.38 \\
MotionLCM & & 0.752 & 0.531 & 9.253 & - & 18.87 & 7.69 & 18.97 \\
TLControl & & 0.779 & 0.271 & 9.569 & - & 0.00 & 0.00 & 1.08 \\
\textbf{Ours} (on pelvis)& & \textcolor{red}{\textbf{0.809}} & \textcolor{red}{\textbf{0.061 (-77\%)}} & \textcolor{red}{\textbf{9.496}} & \textcolor{red}{\textbf{0.0547}} & \textcolor{red}{\textbf{0.00}} & \textcolor{red}{\textbf{0.00}} & \textcolor{red}{\textbf{0.98}} \\
\midrule

\noalign{\vspace{-2pt}}
\multicolumn{9}{c}{\cellcolor{gray!10}\textbf{Train on All Random Joints}} \\
OmniControl &  & ~0.693~ & 0.310 & \textcolor{red}{\textbf{9.502}} & 0.0608 & 6.17 & 1.07 & 4.04 \\
TLControl & Average & 0.782 & 0.256 & 9.719 & - & 0.00 & 0.00 & 1.11 \\
\textbf{Ours} & & \textcolor{red}{\textbf{0.805}} & \textcolor{red}{\textbf{0.083 (-68\%)}} & 9.395 & \textcolor{red}{\textbf{0.0545}} & \textcolor{red}{\textbf{0.00}} & \textcolor{red}{\textbf{0.00}} & \textcolor{red}{\textbf{0.72}} \\
\bottomrule
\end{tabular}}
\vspace{-10pt}
\end{table*}

\textbf{Datasets.} We conduct comprehensive experiments on the HumanML3D dataset \citep{t2m} HumanML3D covers a wide variety motions. It includes 14,616 motion sequences accompanied by 44,970 text descriptions. The textual data contains 5,371 unique words. The motion sequences are sourced from AMASS \citep{AMASS} and HumanAct12 \citep{HumanAct12}.

\textbf{Evaluation.} We follow the evaluation protocol from OmniControl \citep{OmniControl} which combines evaluation of quality from HumanML3D\citep{t2m} and trajectory error from GMD \citep{GMD}. 


\subsection{Quantitative Comparison to State-of-the-art}

GMD \citep{GMD} only addresses the pelvis location on the ground plane (xz coordinates). To ensure a fair comparison, we follow OmniControl \citep{OmniControl} and compare GMD in managing the full 3D position of the pelvis (xyz coordinates).
The first section of Tab. \ref{tab:main_eval} resents results for models trained on the pelvis alone to ensure a fair comparison with previous state-of-the-art methods on the HumanML3D \citep{t2m} dataset. $\rightarrow$ means closer to real data is better. Our model demonstrates significant improvements across all evaluation metrics. When compare to TLControl, the FID score notably decreased from 0.271 to 0.061, the R-Precision increased from 0.779 to 0.809, indicating superior generation quality. In terms of joint control accuracy, both Trajectory Error and Location Error dropped to zero, while the average error decreased to 0.91 cm, indicating highly precise joint control. Furthermore, our model outperforms existing methods in both Diversity and Foot Skating Ratio metrics.
In the second section, \textit{Train on All Joints}, we follow the evaluation from OmniControl \citep{OmniControl}, as our model supports control of any joint, not just the root (pelvis). We train the model to control multiple joints, specifically the pelvis, left foot, right foot, head, left wrist, and right wrist. Our model constantly outperform SOTA in all metrics.


\subsection{Qualitative Comparison to State-of-the-art}
\textbf{Any-joint-Any-Frame.} We visualize the generated motion using GMD \citep{GMD}, MotionLCM \cite{motionlcm} and OmniControl \citep{OmniControl} in Fig. \ref{fig:compare_sota}. The motion is generated based on the prompt ``a person walks forward and waves his hands," with the pelvis and right wrist controlled in a zigzag pattern. Since GMD can only control the pelvis, we apply control only to the pelvis for GMD. However, it fails to follow the zigzag pattern, tending instead to move in a straight line. OmniControl receives control signals for both the pelvis and right wrist. Yet, it not only fails to follow the root trajectory (pelvis) but also does not adhere to the zigzag pattern for the right wrist. In contrast, our MaskControl demonstrates realistic motion with precise joint control for both the pelvis and the right wrist, accurately following the intended zigzag pattern.

\setlength{\parindent}{0pt}
\textbf{Zero-shot Objective Control.} Fig. \ref{fig:zeroshot-sota} shows the visualization results with the unseen objective function, `Constrain a human to walk inside a square area,' compared to ProgMoGen \cite{programmable}. ProgMoGen not only violates the constraint but also generates motion unrelated to the text, as it lacks any `turns around' motion.

\subsection{Body Part Editing}

\begin{table}[!h]
\centering

\vspace{-10pt}
\caption{Quantitative result of upper body editing task on HumanML3D dataset.}
\label{tab:upperbody_editing}
\vspace{-5pt}
\scalebox{0.75}{
\begin{tabular}{lcccccc}
\toprule
\multirow{2}{*}{\textbf{Method}} & \multicolumn{3}{c}{\textbf{R-precision $\uparrow$}} & \multirow{2}{*}{\textbf{\makecell{FID \\$\downarrow$}}} & \multirow{2}{*}{\textbf{\makecell{MM-Dist \\$\downarrow$}}} & \multirow{2}{*}{\textbf{\makecell{Diversity \\ $\rightarrow$}}} \\ 
\cmidrule(lr){2-4}
& \textbf{Top1} & \textbf{Top2} & \textbf{Top3} & & & \\
\midrule
MDM   \citep{MDM}        & 0.298   & 0.462   & 0.571   & 4.827  & 4.598   & 7.010   \\
OmniControl  \citep{OmniControl} & 0.374	& 0.550 & 0.656	& 1.213 &	5.228	& 9.258 \\
MMM   \citep{MMM}        & 0.500    & 0.694  & 0.798  & 0.103 & 2.972   & 9.254  \\
MotionLCM   \citep{motionlcm}        & 0.512    & 0.685  & 0.798  & 0.311 & 2.948   & 9.736  \\
\textbf{Ours}    & \textcolor{red}{\textbf{0.517}}&	\textcolor{red}{\textbf{0.708}}	& \textcolor{red}{\textbf{0.804}}	& \textcolor{red}{\textbf{0.074}}	& \textcolor{red}{\textbf{2.945}}	& \textcolor{red}{\textbf{9.380}} \\ \bottomrule
\end{tabular}}
\vspace{-5pt}
\end{table}

With joint signal controls, our model is capable of conditioning on multiple joints, which can be treated as distinct body parts, while generating the remaining body parts based on text input. In Tab. \ref{tab:upperbody_editing} We quantitatively compare our approach to existing methods designed for this task, including MDM \citep{MDM} and MMM \citep{MMM}. Additionally, we compare it with OmniControl \citep{OmniControl} and MotionLCM \cite{motionlcm}, which also supports joint signal control. However, our evaluation demonstrates that OmniControl performs poorly in this task. Following the evaluation protocol from \citep{MMM}, we condition the lower body parts on ground truth for all frames and generate the upper body based on text descriptions using the HumanML3D dataset \citep{t2m}. Our model is evaluated without retraining, using the same model as in the \textit{Train on All Joints} setup, ensuring a fair comparison with OmniControl, which is trained on a subset of joints. Specifically, we condition only on the pelvis, left foot, and right foot as the lower body signals.

The results show that MDM struggles significantly when conditioned on multiple joints, with the FID score increasing to 4.827. Although OmniControl supports multiple joint control, our experiments reveal that it also suffers under these conditions, with its FID score rising to 1.213. This is consistent with the Cross-Joint evaluation in Tab. \ref{tab:all_evall}, which evaluate on multiple joint combination, where OmniControl's FID score deteriorates considerably. MMM performs well in this task but requires retraining with separate codebooks for upper and lower body parts. In contrast, our model outperforms all other methods across all metrics without any retraining. When comparing to the `Train on Pelvis Only' setup in Tab. \ref{tab:main_eval}, our model achieves similar FID and R-Precision scores, highlighting its robustness in handling multiple joint control signals.

\begin{table*}[]
\centering
\caption{Comparison of zero-shot objective control. Three \textit{Human-Scene Interaction} objectives are adopted from the programmable motion model (ProgMoGen \cite{programmable}). Both ProgMoGen and MaskControl are able to control motion during inference by arbitrary loss functions, while MDM and MoMask serve as uncontrollable baseline models.}
\label{tab:arbitary1}
\vspace{-5pt}
\scalebox{.809}{
\begin{tabular}{l|ccccccc}

\multicolumn{8}{c}{\textbf{Task HSI-1: Head Height Constraint}} \\
\midrule
\textbf{Method} & \textbf{Foot Skate} $\downarrow$ & \textbf{Max Acc.} $\downarrow$ & \textbf{Constraint Error} $\downarrow$ & \textbf{Unsucc. Rate} $\downarrow$ & \textbf{FID} $\downarrow$ & \textbf{Diversity} $\rightarrow$ & \textbf{R-prec. (Top3)} $\uparrow$ \\

\midrule
\rowcolor{gray!10}  MDM \cite{MDM} (Unconstrained)& 0.086 & 0.097 & 0.118 & 0.718 & 0.545 & 9.656 & 0.610 \\  
\rowcolor{gray!10}  MoMask \cite{momask} (Unconstrained) & 0.067 & 0.072 & 0.092 & 0.601 & 0.354 & 9.505 & 0.683 \\  

\midrule
ProgMoGen \cite{programmable} & 0.075 & 0.094 & 0.012 & 0.088 & 0.556 & \textcolor{red}{\textbf{9.611}}   & 0.597 \\ 
\textbf{Ours} & \textcolor{red}{\textbf{0.066}} & \textcolor{red}{\textbf{0.074}} & \textcolor{red}{\textbf{0.000}} & \textcolor{red}{\textbf{0.000}} & \textcolor{red}{\textbf{0.246}} & 9.393 & \textcolor{red}{\textbf{0.695}} \\
\bottomrule
\end{tabular}}

\vspace{4pt}
\scalebox{.77}{
\begin{tabular}{l|ccc|ccc} 
    \multicolumn{1}{c}{} & \multicolumn{3}{c}{\textbf{Task HSI-2: Avoiding Barrier}} &  \multicolumn{3}{c}{\textbf{Task HSI-3: Walking Inside a Square}}  \\
    \midrule
   \textbf{Method} & \textbf{Foot Skate} $\downarrow$  & \textbf{Max Acceleration} $\downarrow$ &  \textbf{Constraint Error} $\downarrow$ & \textbf{Foot Skate} $\downarrow$  & \textbf{Max Acceleration} $\downarrow$ &  \textbf{Constraint Error} $\downarrow$  \\
   \midrule
  \rowcolor{gray!10}    MDM \cite{MDM} (Unconstrained)& 0.096 & 0.126 & 0.454 & 0.096 & 0.126 & 0.301  \\
   \rowcolor{gray!10}   MoMask \cite{momask} (Unconstrained) & 0.072 & 0.117 & 0.464 & 0.072 & 0.117 & 0.270 \\

    \midrule
   
   ProgMoGen \cite{programmable}            & 0.189 & 0.150 & 0.097 & 0.125 & 0.093 & 0.012   \\
   \textbf{Ours}            & \textcolor{red}{\textbf{0.146}} & \textcolor{red}{\textbf{0.126}} & \textcolor{red}{\textbf{0.000}} & \textcolor{red}{\textbf{0.092}} & \textcolor{red}{\textbf{0.078}} & \textcolor{red}{\textbf{0.000}}   \\
  \bottomrule
\end{tabular}}

\vspace{-10pt}

\end{table*}

\subsection{Zero-shot Objective Control}
We evaluate three zero-shot objective control tasks using three Human-Scene Interaction evaluations from ProgMoGen \cite{programmable}, a programmable motion control model, \textit{i.e.} `Head Height Constraint', `Avoiding Barrier', and `Walking Inside a Square'. MDM \cite{MDM} and MoMask \cite{momask} serve as baselines without any control mechanisms. Since ProgMoGen builds on MDM as its base model. Similarly MoMask is shown as a baseline for the unconstrained Mask Motion Model. Tab. \ref{tab:arbitary1} shows that our MaskControl outperform ProgMoGen in all metrics. \textbf{Max Acc.} represents the maximum acceleration of joints which indicates the amount of joint jittering. \textbf{Constraint Error} measures how much the head height exceeds the specified constraint. \textbf{Unsucc. Rate} represents the rate of motion sequences that violate the height constraint.

\section{Ablation Study}


\subsection{Component Analysis}

\begin{table}[!ht]
\centering
\vspace{-10pt}
\caption{Ablation results of components analysis and different densities of joint control signal.}
\label{tab:ablation_study_all}
\scalebox{0.72}{
\begin{tabular}{lcccccc}
\hline
\textbf{Method} & \textbf{\makecell{R-Prec. \\ Top-3 $\uparrow$}} & \textbf{FID $\downarrow$} & \textbf{\makecell{Foot \\ Skat. $\downarrow$}} & \textbf{\makecell{Traj. \\Err.  $\downarrow$}} & \textbf{\makecell{Loc. \\Err.  $\downarrow$}} & \textbf{\makecell{Avg. \\ Err. $\downarrow$}} \\
\hline
\multicolumn{7}{c}{\cellcolor{gray!10}\textbf{Component Analysis}} \\
No Control & 0.807 & 0.095 & \textcolor{red}{\textbf{0.0527}} & 50.66 & 35.11 & 63.18 \\
w/o Logits Regularizer & 0.795 & 0.142 & 0.0577 & 0.32 & 0.02 & 2.18 \\
w/o Logits Optimization & 0.802 & 0.128 & 0.0594 & 39.14 & 24.00 & 40.41 \\
Full model & \textcolor{red}{\textbf{0.809}} & \textcolor{red}{\textbf{0.061}} & 0.0547 & \textcolor{red}{\textbf{0.00}} & \textcolor{red}{\textbf{0.00}} & \textcolor{red}{\textbf{0.98}} \\
\hline
\hline
\multicolumn{7}{c}{\cellcolor{gray!10}\textbf{Density of Spatial Control Signal}} \\
Density: 1 & 0.804 & 0.077 & 0.0551 & 0.00 & 0.00 & 0.10 \\
Density: 2 & 0.806 & 0.087 & 0.0553 & 0.00 & 0.00 & 0.34 \\
Density: 5 & 0.811 & 0.078 & 0.0553 & 0.00 & 0.00 & 0.98 \\
Density: 25\% & 0.812 & 0.055 & 0.0536 & 0.01 & 0.00 & 1.68 \\
Density: 100\% & 0.814 & 0.054 & 0.0543 & 0.02 & 0.00 & 1.64 \\
\hline
\end{tabular}}
\end{table}

To understand how each component impact the quality and joint control error. We conduct an ablation study on each component in sec. \textit{Component Analysis} of Tab. \ref{tab:ablation_study_all}, using same evaluation as Tab. \ref{tab:main_eval}. Without any control, \textbf{No Control}, the model achieves the highest diversity and the lowest Foot Skating Ratio, indicating strong realism in the generated motion. The FID score is also on par. However, all spatial errors are poor due to the absence of joint control components in the model. \textbf{Without the Logits Regularizer}, the average error remains relatively low, but the FID score is the worst, highlighting the importance of the \textit{Logits Regularizer} for generation quality. \textbf{Without Logits Optimization}, the average error significantly worsens, although the FID remains acceptable, suggesting that the \textit{Logits Regularizer} helps maintain generation quality. In the \textbf{Full model}, both the \textit{Logits Regularizer} and \textit{Logits Optimization} complement each other, improving both generation quality (FID) and control accuracy (Average Error).


\subsection{Density of Joint Control Signal}
\label{sec:density}

In second part of table \ref{tab:ablation_study_all}, we provide a detailed analysis of MaskControl's performance across five different joint control density levels, where the model is trained for pelvis control using the HumanML3D dataset. The values 1, 2, and 5 refer to controlling the motion using exactly 1, 2, or 5 frames. The values 25\% and 100\% indicate the percentage of the total ground-truth motion length for each sample, which can range from 40 to 196 frames depending on the sample. The results show that increasing the joint control improves the quality: the FID score decreases from 0.077 with 1-frame control to 0.054 with full 196-frame (100\%) control. Similarly, R-Precision improves from 0.804 at 1-frame density to 0.814 at 196-frame (100\%) density. However, the Average Error shows the opposite trend—more joint control leads to higher error, as the model is required to target more specific points.

\section{Conclusion}
MaskControl is the first model to introduce controllability to the generative masked motion model, enabling precise control while maintaining high-quality motion generation, consistently outperforming diffusion-based controllable frameworks. MaskControl introduces \textit{Differentiable Expectation Sampling (DES)} to relax the quantization constraint, enabling two key innovations: \textit{Logits Regularizer} uses random masking and reconstruction to ensure that the generated motions are of high fidelity, while also reducing inconsistencies between the input control signals and the motions produced. \textit{Inference-Time Logit Optimization} fine-tunes the predicted motion distribution during the unmasking process to the input control signals, enhancing precision and making MaskControl adaptable for unseen tasks. MaskControl has a wide range of applications, including any-joint-any-frame control, body-part timeline control, and zero-shot objective control.

{
    \small
    \bibliographystyle{ieeenat_fullname}
    \bibliography{ref/diffusion, ref/intro, ref/token,ref/t2m, ref/other}

\begin{thebibliography}{62}
\providecommand{\natexlab}[1]{#1}
\providecommand{\url}[1]{\texttt{#1}}
\expandafter\ifx\csname urlstyle\endcsname\relax
  \providecommand{\doi}[1]{doi: #1}\else
  \providecommand{\doi}{doi: \begingroup \urlstyle{rm}\Url}\fi

\bibitem[Ahuja and Morency(2019)]{Language2Pose}
Chaitanya Ahuja and Louis-Philippe Morency.
\newblock Language2pose: Natural language grounded pose forecasting.
\newblock In \emph{2019 International Conference on 3D Vision (3DV)}, pages 719--728, 2019.

\bibitem[Cha et~al.(2024)Cha, Kim, Yoon, and Baek]{Text2HOI}
Junuk Cha, Jihyeon Kim, Jae~Shin Yoon, and Seungryul Baek.
\newblock Text2hoi: Text-guided 3d motion generation for hand-object interaction.
\newblock In \emph{Proceedings of the IEEE/CVF Conference on Computer Vision and Pattern Recognition}, pages 1577--1585, 2024.

\bibitem[Chang et~al.(2022)Chang, Zhang, Jiang, Liu, and Freeman]{MaskGIT}
Huiwen Chang, Han Zhang, Lu Jiang, Ce Liu, and William~T. Freeman.
\newblock Maskgit: Masked generative image transformer.
\newblock \emph{2022 IEEE/CVF Conference on Computer Vision and Pattern Recognition (CVPR)}, pages 11305--11315, 2022.

\bibitem[Chang et~al.(2023)Chang, Zhang, Barber, Maschinot, Lezama, Jiang, Yang, Murphy, Freeman, Rubinstein, Li, and Krishnan]{Muse}
Huiwen Chang, Han Zhang, Jarred Barber, AJ Maschinot, Jos{\'e} Lezama, Lu Jiang, Ming Yang, Kevin~P. Murphy, William~T. Freeman, Michael Rubinstein, Yuanzhen Li, and Dilip Krishnan.
\newblock Muse: Text-to-image generation via masked generative transformers.
\newblock \emph{ArXiv}, abs/2301.00704, 2023.

\bibitem[Chen et~al.(2022)Chen, Jiang, Liu, Huang, Fu, Chen, Yu, and Yu]{MLD}
Xin Chen, Biao Jiang, Wen Liu, Zilong Huang, Bin Fu, Tao Chen, Jingyi Yu, and Gang Yu.
\newblock Executing your commands via motion diffusion in latent space.
\newblock \emph{2023 IEEE/CVF Conference on Computer Vision and Pattern Recognition (CVPR)}, pages 18000--18010, 2022.

\bibitem[Cohan et~al.(2024)Cohan, Tevet, Reda, Peng, and van~de Panne]{cohan2024flexible}
Setareh Cohan, Guy Tevet, Daniele Reda, Xue~Bin Peng, and Michiel van~de Panne.
\newblock Flexible motion in-betweening with diffusion models.
\newblock In \emph{ACM SIGGRAPH 2024 Conference Papers}, pages 1--9, 2024.

\bibitem[Dai et~al.(2024)Dai, Chen, Wang, Liu, Dai, and Tang]{motionlcm}
Wenxun Dai, Ling-Hao Chen, Jingbo Wang, Jinpeng Liu, Bo Dai, and Yansong Tang.
\newblock Motionlcm: Real-time controllable motion generation via latent consistency model.
\newblock \emph{arXiv preprint arXiv:2404.19759}, 2024.

\bibitem[Devlin et~al.(2019)Devlin, Chang, Lee, and Toutanova]{BERT}
Jacob Devlin, Ming-Wei Chang, Kenton Lee, and Kristina Toutanova.
\newblock Bert: Pre-training of deep bidirectional transformers for language understanding.
\newblock In \emph{North American Chapter of the Association for Computational Linguistics}, 2019.

\bibitem[Diller and Dai(2024)]{CgHoi}
Christian Diller and Angela Dai.
\newblock Cg-hoi: Contact-guided 3d human-object interaction generation.
\newblock In \emph{Proceedings of the IEEE/CVF Conference on Computer Vision and Pattern Recognition}, pages 19888--19901, 2024.

\bibitem[Diomataris et~al.(2024)Diomataris, Athanasiou, Taheri, Wang, Hilliges, and Black]{wandr}
Markos Diomataris, Nikos Athanasiou, Omid Taheri, Xi Wang, Otmar Hilliges, and Michael~J. Black.
\newblock {WANDR}: Intention-guided human motion generation.
\newblock In \emph{Proceedings IEEE Conference on Computer Vision and Pattern Recognition (CVPR)}, 2024.

\bibitem[Du et~al.(2023)Du, Kips, Pumarola, Starke, Thabet, and Sanakoyeu]{AGRoL}
Yuming Du, Robin Kips, Albert Pumarola, Sebastian Starke, Ali Thabet, and Artsiom Sanakoyeu.
\newblock Avatars grow legs: Generating smooth human motion from sparse tracking inputs with diffusion model.
\newblock In \emph{Proceedings of the IEEE/CVF Conference on Computer Vision and Pattern Recognition}, pages 481--490, 2023.

\bibitem[Esser et~al.(2020)Esser, Rombach, and Ommer]{vqgan}
Patrick Esser, Robin Rombach, and Bj{\"o}rn Ommer.
\newblock Taming transformers for high-resolution image synthesis.
\newblock \emph{2021 IEEE/CVF Conference on Computer Vision and Pattern Recognition (CVPR)}, pages 12868--12878, 2020.

\bibitem[Guo et~al.(2020)Guo, Zuo, Wang, Zou, Sun, Deng, Gong, and Cheng]{HumanAct12}
Chuan Guo, Xinxin Zuo, Sen Wang, Shihao Zou, Qingyao Sun, Annan Deng, Minglun Gong, and Li Cheng.
\newblock Action2motion: Conditioned generation of 3d human motions.
\newblock \emph{Proceedings of the 28th ACM International Conference on Multimedia}, 2020.

\bibitem[Guo et~al.(2022{\natexlab{a}})Guo, Xuo, Wang, and Cheng]{TM2T}
Chuan Guo, Xinxin Xuo, Sen Wang, and Li Cheng.
\newblock Tm2t: Stochastic and tokenized modeling for the reciprocal generation of 3d human motions and texts.
\newblock \emph{ArXiv}, abs/2207.01696, 2022{\natexlab{a}}.

\bibitem[Guo et~al.(2022{\natexlab{b}})Guo, Zou, Zuo, Wang, Ji, Li, and Cheng]{t2m}
Chuan Guo, Shihao Zou, Xinxin Zuo, Sen Wang, Wei Ji, Xingyu Li, and Li Cheng.
\newblock Generating diverse and natural 3d human motions from text.
\newblock In \emph{2022 IEEE/CVF Conference on Computer Vision and Pattern Recognition (CVPR)}, pages 5142--5151, 2022{\natexlab{b}}.

\bibitem[Guo et~al.(2023)Guo, Mu, Javed, Wang, and Cheng]{momask}
Chuan Guo, Yuxuan Mu, Muhammad~Gohar Javed, Sen Wang, and Li Cheng.
\newblock Momask: Generative masked modeling of 3d human motions.
\newblock 2023.

\bibitem[Huang et~al.(2023)Huang, Wang, Li, Jia, Liu, Zhu, Liang, and Zhu]{SceneDiffuser}
Siyuan Huang, Zan Wang, Puhao Li, Baoxiong Jia, Tengyu Liu, Yixin Zhu, Wei Liang, and Song-Chun Zhu.
\newblock Diffusion-based generation, optimization, and planning in 3d scenes.
\newblock In \emph{Proceedings of the IEEE/CVF Conference on Computer Vision and Pattern Recognition}, pages 16750--16761, 2023.

\bibitem[Huang et~al.(2024)Huang, Wan, Yang, Callison-Burch, Yatskar, and Liu]{CoMo}
Yiming Huang, Weilin Wan, Yue Yang, Chris Callison-Burch, Mark Yatskar, and Lingjie Liu.
\newblock Como: Controllable motion generation through language guided pose code editing, 2024.

\bibitem[Jang et~al.(2017)Jang, Gu, and Poole]{gumbelsoftmax}
Eric Jang, Shixiang Gu, and Ben Poole.
\newblock Categorical reparametrization with gumble-softmax.
\newblock In \emph{International Conference on Learning Representations (ICLR 2017)}. OpenReview. net, 2017.

\bibitem[Jiang et~al.(2023)Jiang, Chen, Liu, Yu, Yu, and Chen]{MotionGPT}
Biao Jiang, Xin Chen, Wen Liu, Jingyi Yu, Gang Yu, and Tao Chen.
\newblock Motiongpt: Human motion as a foreign language.
\newblock \emph{ArXiv}, abs/2306.14795, 2023.

\bibitem[Karunratanakul et~al.(2023)Karunratanakul, Preechakul, Suwajanakorn, and Tang]{GMD}
Korrawe Karunratanakul, Konpat Preechakul, Supasorn Suwajanakorn, and Siyu Tang.
\newblock Guided motion diffusion for controllable human motion synthesis.
\newblock In \emph{Proceedings of the IEEE/CVF International Conference on Computer Vision}, pages 2151--2162, 2023.

\bibitem[Karunratanakul et~al.(2024)Karunratanakul, Preechakul, Aksan, Beeler, Suwajanakorn, and Tang]{DNO}
Korrawe Karunratanakul, Konpat Preechakul, Emre Aksan, Thabo Beeler, Supasorn Suwajanakorn, and Siyu Tang.
\newblock Optimizing diffusion noise can serve as universal motion priors.
\newblock In \emph{Proceedings of the IEEE/CVF Conference on Computer Vision and Pattern Recognition}, pages 1334--1345, 2024.

\bibitem[Kim et~al.(2022)Kim, Kim, and Choi]{FLAME}
Jihoon Kim, Jiseob Kim, and Sungjoon Choi.
\newblock Flame: Free-form language-based motion synthesis \& editing.
\newblock In \emph{AAAI Conference on Artificial Intelligence}, 2022.

\bibitem[Kong et~al.(2023)Kong, Gong, Lian, Mi, and Wang]{kong2023priority}
Hanyang Kong, Kehong Gong, Dongze Lian, Michael~Bi Mi, and Xinchao Wang.
\newblock Priority-centric human motion generation in discrete latent space.
\newblock In \emph{Proceedings of the IEEE/CVF International Conference on Computer Vision}, pages 14806--14816, 2023.

\bibitem[Kulkarni et~al.(2024)Kulkarni, Rempe, Genova, Kundu, Johnson, Fouhey, and Guibas]{Nifty}
Nilesh Kulkarni, Davis Rempe, Kyle Genova, Abhijit Kundu, Justin Johnson, David Fouhey, and Leonidas Guibas.
\newblock Nifty: Neural object interaction fields for guided human motion synthesis.
\newblock In \emph{Proceedings of the IEEE/CVF Conference on Computer Vision and Pattern Recognition}, pages 947--957, 2024.

\bibitem[Lee et~al.(2019)Lee, Yang, Liu, Wang, Lu, Yang, and Kautz]{Dancingtomusic}
Hsin-Ying Lee, Xiaodong Yang, Ming-Yu Liu, Ting-Chun Wang, Yu-Ding Lu, Ming-Hsuan Yang, and Jan Kautz.
\newblock Dancing to music.
\newblock In \emph{Advances in Neural Information Processing Systems}. Curran Associates, Inc., 2019.

\bibitem[Li et~al.(2021{\natexlab{a}})Li, Zhao, Shi, and Sheng]{DanceFormer}
Buyu Li, Yongchi Zhao, Zhelun Shi, and Lu Sheng.
\newblock Danceformer: Music conditioned 3d dance generation with parametric motion transformer.
\newblock In \emph{AAAI Conference on Artificial Intelligence}, 2021{\natexlab{a}}.

\bibitem[Li et~al.(2023)Li, Clegg, Mottaghi, Wu, Puig, and Liu]{CHOIS}
Jiaman Li, Alexander Clegg, Roozbeh Mottaghi, Jiajun Wu, Xavier Puig, and C~Karen Liu.
\newblock Controllable human-object interaction synthesis.
\newblock \emph{arXiv preprint arXiv:2312.03913}, 2023.

\bibitem[Li et~al.(2021{\natexlab{b}})Li, Yang, Ross, and Kanazawa]{AIChoreographer}
Ruilong Li, Sha Yang, David~A. Ross, and Angjoo Kanazawa.
\newblock Ai choreographer: Music conditioned 3d dance generation with aist++.
\newblock \emph{2021 IEEE/CVF International Conference on Computer Vision (ICCV)}, pages 13381--13392, 2021{\natexlab{b}}.

\bibitem[Liu et~al.(2024)Liu, Zhan, Huang, Mu, and Shan]{programmable}
Hanchao Liu, Xiaohang Zhan, Shaoli Huang, Tai-Jiang Mu, and Ying Shan.
\newblock Programmable motion generation for open-set motion control tasks.
\newblock \emph{CVPR}, 2024.

\bibitem[Lou et~al.(2023)Lou, Zhu, Wang, Wang, and Yang]{DiverseMotion}
Yunhong Lou, Linchao Zhu, Yaxiong Wang, Xiaohan Wang, and Yezhou Yang.
\newblock Diversemotion: Towards diverse human motion generation via discrete diffusion.
\newblock \emph{ArXiv}, abs/2309.01372, 2023.

\bibitem[Luo et~al.(2023{\natexlab{a}})Luo, Cao, Kitani, Xu, et~al.]{PHC}
Zhengyi Luo, Jinkun Cao, Kris Kitani, Weipeng Xu, et~al.
\newblock Perpetual humanoid control for real-time simulated avatars.
\newblock In \emph{Proceedings of the IEEE/CVF International Conference on Computer Vision}, pages 10895--10904, 2023{\natexlab{a}}.

\bibitem[Luo et~al.(2023{\natexlab{b}})Luo, Cao, Merel, Winkler, Huang, Kitani, and Xu]{pulse}
Zhengyi Luo, Jinkun Cao, Josh Merel, Alexander Winkler, Jing Huang, Kris Kitani, and Weipeng Xu.
\newblock Universal humanoid motion representations for physics-based control.
\newblock \emph{arXiv preprint arXiv:2310.04582}, 2023{\natexlab{b}}.

\bibitem[Mahmood et~al.(2019)Mahmood, Ghorbani, Troje, Pons-Moll, and Black]{AMASS}
Naureen Mahmood, Nima Ghorbani, Nikolaus~F. Troje, Gerard Pons-Moll, and Michael~J. Black.
\newblock Amass: Archive of motion capture as surface shapes.
\newblock \emph{2019 IEEE/CVF International Conference on Computer Vision (ICCV)}, pages 5441--5450, 2019.

\bibitem[Peng et~al.(2021)Peng, Ma, Abbeel, Levine, and Kanazawa]{AMP}
Xue~Bin Peng, Ze Ma, P. Abbeel, Sergey Levine, and Angjoo Kanazawa.
\newblock Amp.
\newblock \emph{ACM Transactions on Graphics (TOG)}, 40:\penalty0 1 -- 20, 2021.

\bibitem[Peng et~al.(2022)Peng, Guo, Halper, Levine, and Fidler]{ASE}
Xue~Bin Peng, Yunrong Guo, Lina Halper, Sergey Levine, and Sanja Fidler.
\newblock Ase.
\newblock \emph{ACM Transactions on Graphics (TOG)}, 41:\penalty0 1 -- 17, 2022.

\bibitem[Petrovich et~al.(2022)Petrovich, Black, and Varol]{TEMOS}
Mathis Petrovich, Michael~J. Black, and G{\"u}l Varol.
\newblock Temos: Generating diverse human motions from textual descriptions.
\newblock \emph{ArXiv}, abs/2204.14109, 2022.

\bibitem[Petrovich et~al.(2024)Petrovich, Litany, Iqbal, Black, Varol, Peng, and Rempe]{Multi-TrackTimelineControl}
Mathis Petrovich, Or Litany, Umar Iqbal, Michael~J. Black, G{\"u}l Varol, Xue~Bin Peng, and Davis Rempe.
\newblock Multi-track timeline control for text-driven 3d human motion generation.
\newblock In \emph{CVPR Workshop on Human Motion Generation}, 2024.

\bibitem[Pinyoanuntapong et~al.(2024{\natexlab{a}})Pinyoanuntapong, Saleem, Wang, Lee, Das, and Chen]{BAMM}
Ekkasit Pinyoanuntapong, Muhammad~Usama Saleem, Pu Wang, Minwoo Lee, Srijan Das, and Chen Chen.
\newblock Bamm: Bidirectional autoregressive motion model.
\newblock In \emph{Computer Vision -- ECCV 2024}, 2024{\natexlab{a}}.

\bibitem[Pinyoanuntapong et~al.(2024{\natexlab{b}})Pinyoanuntapong, Wang, Lee, and Chen]{MMM}
Ekkasit Pinyoanuntapong, Pu Wang, Minwoo Lee, and Chen Chen.
\newblock Mmm: Generative masked motion model.
\newblock In \emph{Proceedings of the IEEE/CVF Conference on Computer Vision and Pattern Recognition (CVPR)}, 2024{\natexlab{b}}.

\bibitem[Radford et~al.(2021)Radford, Kim, Hallacy, Ramesh, Goh, Agarwal, Sastry, Askell, Mishkin, Clark, Krueger, and Sutskever]{CLIP}
Alec Radford, Jong~Wook Kim, Chris Hallacy, Aditya Ramesh, Gabriel Goh, Sandhini Agarwal, Girish Sastry, Amanda Askell, Pamela Mishkin, Jack Clark, Gretchen Krueger, and Ilya Sutskever.
\newblock Learning transferable visual models from natural language supervision.
\newblock In \emph{International Conference on Machine Learning}, 2021.

\bibitem[Rempe et~al.(2023)Rempe, Luo, Peng, Yuan, Kitani, Kreis, Fidler, and Litany]{trace}
Davis Rempe, Zhengyi Luo, Xue~Bin Peng, Ye Yuan, Kris Kitani, Karsten Kreis, Sanja Fidler, and Or Litany.
\newblock Trace and pace: Controllable pedestrian animation via guided trajectory diffusion.
\newblock In \emph{Conference on Computer Vision and Pattern Recognition (CVPR)}, 2023.

\bibitem[Shafir et~al.(2023)Shafir, Tevet, Kapon, and Bermano]{PriorMDM}
Yonatan Shafir, Guy Tevet, Roy Kapon, and Amit~H. Bermano.
\newblock Human motion diffusion as a generative prior.
\newblock \emph{ArXiv}, abs/2303.01418, 2023.

\bibitem[Siyao et~al.(2022)Siyao, Yu, Gu, Lin, Wang, Qian, Loy, and Liu]{Bailando}
Lian Siyao, Weijiang Yu, Tianpei Gu, Chunze Lin, Quan Wang, Chen Qian, Chen~Change Loy, and Ziwei Liu.
\newblock Bailando: 3d dance generation by actor-critic gpt with choreographic memory.
\newblock \emph{2022 IEEE/CVF Conference on Computer Vision and Pattern Recognition (CVPR)}, pages 11040--11049, 2022.

\bibitem[Siyao et~al.(2023)Siyao, Yu, Gu, Lin, Wang, Qian, Loy, and Liu]{Bailando++}
Li Siyao, Weijiang Yu, Tianpei Gu, Chunze Lin, Quan Wang, Chen Qian, Chen~Change Loy, and Ziwei Liu.
\newblock Bailando++: 3d dance gpt with choreographic memory.
\newblock \emph{IEEE Transactions on Pattern Analysis and Machine Intelligence}, pages 1--15, 2023.

\bibitem[Tessler et~al.(2024)Tessler, Guo, Nabati, Chechik, and Peng]{maskedmimic}
Chen Tessler, Yunrong Guo, Ofir Nabati, Gal Chechik, and Xue~Bin Peng.
\newblock Maskedmimic: Unified physics-based character control through masked motion inpainting.
\newblock \emph{ACM Transactions on Graphics (TOG)}, 2024.

\bibitem[Tevet et~al.(2022{\natexlab{a}})Tevet, Gordon, Hertz, Bermano, and Cohen-Or]{MotionCLIP}
Guy Tevet, Brian Gordon, Amir Hertz, Amit~H. Bermano, and Daniel Cohen-Or.
\newblock Motionclip: Exposing human motion generation to clip space.
\newblock In \emph{European Conference on Computer Vision}, 2022{\natexlab{a}}.

\bibitem[Tevet et~al.(2022{\natexlab{b}})Tevet, Raab, Gordon, Shafir, Cohen-Or, and Bermano]{MDM}
Guy Tevet, Sigal Raab, Brian Gordon, Yonatan Shafir, Daniel Cohen-Or, and Amit~H. Bermano.
\newblock Human motion diffusion model.
\newblock \emph{ArXiv}, abs/2209.14916, 2022{\natexlab{b}}.

\bibitem[Tseng et~al.(2022)Tseng, Castellon, and Liu]{EDGE}
Jo-Han Tseng, Rodrigo Castellon, and C.~Karen Liu.
\newblock Edge: Editable dance generation from music.
\newblock \emph{2023 IEEE/CVF Conference on Computer Vision and Pattern Recognition (CVPR)}, pages 448--458, 2022.

\bibitem[Villegas et~al.(2022)Villegas, Babaeizadeh, Kindermans, Moraldo, Zhang, Saffar, Castro, Kunze, and Erhan]{Phenaki}
Ruben Villegas, Mohammad Babaeizadeh, Pieter-Jan Kindermans, Hernan Moraldo, Han Zhang, Mohammad~Taghi Saffar, Santiago Castro, Julius Kunze, and D. Erhan.
\newblock Phenaki: Variable length video generation from open domain textual description.
\newblock \emph{ArXiv}, abs/2210.02399, 2022.

\bibitem[Wan et~al.(2023)Wan, Dou, Komura, Wang, Jayaraman, and Liu]{TLcontrol}
Weilin Wan, Zhiyang Dou, Taku Komura, Wenping Wang, Dinesh Jayaraman, and Lingjie Liu.
\newblock Tlcontrol: Trajectory and language control for human motion synthesis.
\newblock \emph{arXiv preprint arXiv:2311.17135}, 2023.

\bibitem[Wang et~al.(2024)Wang, Chen, Jia, Li, Zhang, Zhang, Liu, Zhu, Liang, and Huang]{MoveasYouSay}
Zan Wang, Yixin Chen, Baoxiong Jia, Puhao Li, Jinlu Zhang, Jingze Zhang, Tengyu Liu, Yixin Zhu, Wei Liang, and Siyuan Huang.
\newblock Move as you say, interact as you can: Language-guided human motion generation with scene affordance.
\newblock In \emph{Proceedings of the IEEE/CVF Conference on Computer Vision and Pattern Recognition (CVPR)}, 2024.

\bibitem[Williams et~al.(2020)Williams, Ringer, Ash, Hughes, Macleod, and Dougherty]{Hierarchical-vqvae}
Will Williams, Sam Ringer, Tom Ash, John Hughes, David Macleod, and Jamie Dougherty.
\newblock Hierarchical quantized autoencoders.
\newblock \emph{ArXiv}, abs/2002.08111, 2020.

\bibitem[Xie et~al.(2021)Xie, Wang, Iqbal, Guo, Fidler, and Shkurti]{PhysicsbasedHM}
Kevin Xie, Tingwu Wang, Umar Iqbal, Yunrong Guo, Sanja Fidler, and Florian Shkurti.
\newblock Physics-based human motion estimation and synthesis from videos.
\newblock \emph{2021 IEEE/CVF International Conference on Computer Vision (ICCV)}, pages 11512--11521, 2021.

\bibitem[Xie et~al.(2023)Xie, Jampani, Zhong, Sun, and Jiang]{OmniControl}
Yiming Xie, Varun Jampani, Lei Zhong, Deqing Sun, and Huaizu Jiang.
\newblock Omnicontrol: Control any joint at any time for human motion generation.
\newblock 2023.

\bibitem[Yan et~al.(2023)Yan, Liu, Wang, Du, Liu, and Liu]{CrossModalRF}
Sheng Yan, Yang Liu, Haoqiang Wang, Xin Du, Mengyuan Liu, and Hong Liu.
\newblock Cross-modal retrieval for motion and text via doptriple loss.
\newblock 2023.

\bibitem[Yuan et~al.(2022)Yuan, Song, Iqbal, Vahdat, and Kautz]{PhysDiff}
Ye Yuan, Jiaming Song, Umar Iqbal, Arash Vahdat, and Jan Kautz.
\newblock Physdiff: Physics-guided human motion diffusion model.
\newblock \emph{ArXiv}, abs/2212.02500, 2022.

\bibitem[Zhang et~al.(2023{\natexlab{a}})Zhang, Zhang, Cun, Huang, Zhang, Zhao, Lu, and Shen]{T2M-GPT}
Jianrong Zhang, Yangsong Zhang, Xiaodong Cun, Shaoli Huang, Yong Zhang, Hongwei Zhao, Hongtao Lu, and Xiaodong Shen.
\newblock Generating human motion from textual descriptions with discrete representations.
\newblock \emph{2023 IEEE/CVF Conference on Computer Vision and Pattern Recognition (CVPR)}, pages 14730--14740, 2023{\natexlab{a}}.

\bibitem[Zhang et~al.(2023{\natexlab{b}})Zhang, Rao, and Agrawala]{ControlNet}
Lvmin Zhang, Anyi Rao, and Maneesh Agrawala.
\newblock Adding conditional control to text-to-image diffusion models.
\newblock In \emph{Proceedings of the IEEE/CVF International Conference on Computer Vision}, pages 3836--3847, 2023{\natexlab{b}}.

\bibitem[Zhang et~al.(2022)Zhang, Cai, Pan, Hong, Guo, Yang, and Liu]{MotionDiffuse}
Mingyuan Zhang, Zhongang Cai, Liang Pan, Fangzhou Hong, Xinying Guo, Lei Yang, and Ziwei Liu.
\newblock Motiondiffuse: Text-driven human motion generation with diffusion model.
\newblock \emph{ArXiv}, abs/2208.15001, 2022.

\bibitem[Zhong et~al.(2023)Zhong, Hu, Zhang, and Xia]{AttT2M}
Chongyang Zhong, Lei Hu, Zihao Zhang, and Shihong Xia.
\newblock Attt2m: Text-driven human motion generation with multi-perspective attention mechanism.
\newblock \emph{ArXiv}, abs/2309.00796, 2023.

\bibitem[Zhong et~al.(2024)Zhong, Xie, Jampani, Sun, and Jiang]{SMooDi}
Lei Zhong, Yiming Xie, Varun Jampani, Deqing Sun, and Huaizu Jiang.
\newblock Smoodi: Stylized motion diffusion model.
\newblock \emph{arXiv preprint arXiv:2407.12783}, 2024.

\end{thebibliography}
}

\clearpage
\appendix
\onecolumn
\section{Appendix}
\subsection{Overview}
\label{sec:Summary}
The supplementary material is organized into the following sections:
\begin{itemize}
    \item Section \ref{sec:STMC}: Comparision to STMC
    \item Section \ref{sec:algorithm}: Pseudo Code of MaskControl Inference
    \item Section \ref{sec:implementation}: Implementation Details
    \item Section \ref{sec:eval_all_each}: Full Evaluation on All Joint
    \item Section \ref{sec:speed_quality_error}: Inference speed, quality, and errors Details
    \item Section \ref{sec:speed_component}: Speed of each component
    \item Section \ref{sec:quantitative_result_fast}: Quantitative result for all joints of MaskControl-Fast
    \item Section \ref{sec:lesser_iteration}: Ablation on less number of generation steps
    \item Section \ref{sec:analysis_of_logits}: Analysis of \textit{Logits Optimization} and \textit{Logits Regularizer}
    \item Section \ref{sec:challenges_motioncontrol}: Ambiguity of Motion Control Signal
    \item Section \ref{sec:bodypart}: Body Part Timeline Control
    \item Section \ref{sec:cross_combination}: Cross Combination
\end{itemize}
Video visualization can be found at \url{https://www.ekkasit.com/ControlMM-page/}

\subsection{Comparision to STMC}
\label{sec:STMC}
The STMC \cite{Multi-TrackTimelineControl} setting involves different controllable joints and longer motion sequences, making direct comparison infeasible without re-training the \textit{Logits Regularizer}. Fortunately, \textit{Logits Optimization} enables zero-shot objective control, allowing us to achieve the STMC setting without retraining. We adopt the following steps: \textbf{(1) Generate motions for each prompt}: Motions are generated separately without joint control. \textbf{(2) Resolving unassigned timeframes}: Following STMC, we apply the SINC heuristic to fill in unassigned joints based on the sparse body-part timeline input. \textbf{(3) Any-Joint-Any-Frame Control by each body part}: We use the control conditions from steps (1) and (2), and remove joint control from connected timeframes to create padding, enabling smooth transitions between adjacent motion segments. Finally, the full motion is generated using these joint control signals without any text prompt, effectively generating body-part timelines onto a blank canvas.

\begin{figure}[ht]
 \centering
  \includegraphics[width=.8\textwidth]{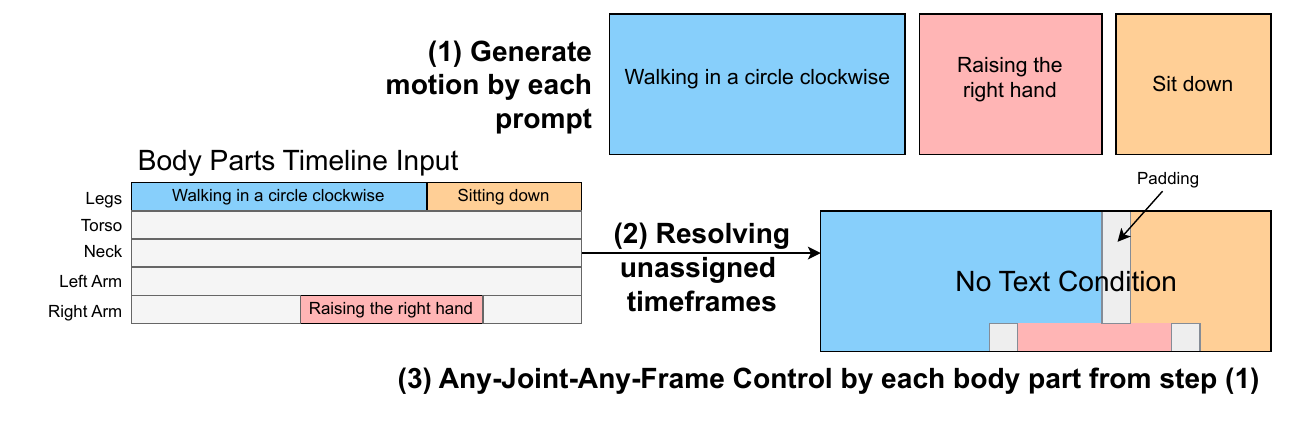}
  \caption{Generating body parts timeline for STMC setting.}
  \label{fig:bamm_compare}
\end{figure}



\noindent We evaluate two variants: \textbf{`No Control Pad'}, where no control is applied in the padding regions, and \textbf{`Merge Control Pad'}, where the optimization weight gradually decreases from adjacent body-part controls. Our method outperforms STMC across all metrics, as shown in Tab. \ref{tab:STMC_eval}. Note that performance could be further improved by re-training the \textit{Logits Regularizer} for the STMC setting.

\begin{table}[h]
    \centering
    \setlength{\tabcolsep}{6pt}
    \resizebox{.7\linewidth}{!}{
    \begin{tabular}{l|cccc|cc}
        \toprule
        & \multicolumn{4}{c|}{Per-crop semantic correctness} & \multicolumn{2}{c}{Realism} \\
        Method & \multirow{2}{*}{R@1 $\uparrow$} & \multirow{2}{*}{R@3 $\uparrow$} & \multicolumn{2}{c|}{TMR-Score $\uparrow$} & \multirow{2}{*}{FID $\downarrow$} & Transition \\
        & & & M2T & M2M & & distance $\downarrow$ \\
        \midrule
        \textbf{Ground truth}     & 55.0 & 73.3 & 0.748 & 1.000 & 0.000 & 1.5 \\
        \midrule
        STMC-MotionDiffuse    & 24.8 & 46.7 & 0.660 & 0.632 & 0.531 & \textcolor{blue}{\textbf{1.5}} \\
        STMC-MDM              & 25.1 & 46.0 & 0.641 & 0.633 & 0.606 & 2.4 \\
        \textbf{Our (No Control Pad)}& \textcolor{red}{\textbf{38.3}} & \textcolor{red}{\textbf{58.1}} & \textcolor{red}{\textbf{0.688}} & \textcolor{red}{\textbf{0.654}} & \textcolor{blue}{\textbf{0.511}} & 1.6 \\
        \textbf{Our (Merge Control Pad)} &  \textcolor{blue}{\textbf{34.8}} & \textcolor{blue}{\textbf{55.6}} &  \textcolor{blue}{\textbf{0.675}} &  \textcolor{blue}{\textbf{0.653}} & \textcolor{red}{\textbf{0.508}} &  \textcolor{red}{\textbf{1.5}} \\
        \bottomrule        
    \end{tabular}
    }
    \caption{Quantitative comparison with STMC}
    \label{tab:STMC_eval}
\end{table}

\subsection{Pseudo Code of MaskControl Inference}
\label{sec:algorithm}

\begin{algorithm}[H]
\caption{MaskControl Inference}
\label{alg:inference}
\begin{algorithmic}[1]
\Require Masked Motion Model ($MMM$),  Logits Regularizer ($LR$), mask scheduling function $\gamma(\cdot)$, spatial control signals $s$ (if any), text prompts $W$ (if any).
\State $X_{\overline{\mathbf{M}}} \leftarrow \textit{[Mask]}$ \Comment{\small Start with all mask tokens}
\ForAll{$t$ from $1$ to $T$}    \Comment{\small Unmask process in $T$ steps}
    \State $\{\vf\} \leftarrow LR(X_{\overline{\mathbf{M}}}, W, s; \phi)$ \Comment{\small \textbf{Logits Regularizer}}
    \State $l \leftarrow MMM(X_{\overline{\mathbf{M}}}, W, \{\vf\};\theta)$ \Comment{\small Masked Motion Model}
    \ForAll{$i$ from $1$ to $I_l$} \Comment{\small \textbf{Logits Optimization}}
        \State $l_{i+1} = l_i - \eta \nabla_{l_i} L_s(l_i, s) $
    \EndFor
    \State $X_{\overline{\mathbf{M}}} \leftarrow \gamma(l, t)$ \Comment{\small mask out tokens based on logits $l$ at time step $t$}
\EndFor
\ForAll{$i$ from $1$ to $I_e$} \Comment{\small \textbf{Logits Optimization}}
    \State $e_c^{i+1} = e_c^i - \eta \nabla_{e_c^i} L_s(e_c^i, s)$
\EndFor

\\
\Return $Decoder(e_c)$
\end{algorithmic}
\end{algorithm}

\subsection{Implementation Details}
\label{sec:implementation}
We modified the MoMask \citep{momask} model by retraining it with a cross-entropy loss applied to all tokens, instead of just the masked positions. This retrained model serves as our pretrained base model, and we kept the default hyperparameter settings unchanged. To improve robustness to text variation, we randomly drop 10\% of the text conditioning, which also allows the model to be used for Classifier-Free Guidance (CFG). The weight for Eq. \ref{eq:all_controll_loss} is set to $\alpha = 0.1$. We use a codebook of size 512, with embeddings of size 512 and 6 residual layers. The Transformer embedding size is set to 384, with 6 attention heads, each with an embedding size of 64, distributed across 8 layers. This configuration demonstrates the feasibility of converting between two different embedding sizes and spaces using the Differentiable Expectation Sampling 
(DES). The encoder and decoder downsample the motion sequence length by a factor of 4 when mapping to token space. The learning rate follows a linear warm-up schedule, reaching 2e-4 after 2000 iterations, using AdamW optimization. The mini-batch size is set to 512 for training RVQ-VAE and 64 for training the Transformers. During inference, the CFG scale is set to $cfg = 4$ for the base layer and $cfg = 5$ for the 6 layers of residual, with 10 steps for generation. We use pretrained CLIP model \citep{CLIP} to generate text embeddings, which have a size of 512. These embeddings are then projected down to a size of 384 to match the token size used by the Transformer. \textit{Logits Regularizer} is a trainable copy of Masked Transformer with the zero linear layer connect to the output each layer of the Masked Transformer. During inference, \textit{Logits Optimization} applies L2 loss with a learning rate of $0.06$ for 100 iterations in \textit{Logits Optimization} for each of the 10 generation steps and 600 iterations in the last unmasking step. We apply temperature of 1 for all 10 steps and 1e-8 for residual layers. We follow the implementation from \cite{GMD, OmniControl, TLcontrol}, applying the spatial control signal only to joint positions and omitting rotations.

\subsection{Full Evaluation on All Joint}
\label{sec:eval_all_each}
Following the evaluation from OmniControl \cite{OmniControl}, Table \ref{tab:all_evall} extends Table \ref{tab:main_eval} by showing the evaluation for each joint individually. Our MaskControl outperforms SOTA across all metrics. `Cross' is the random combination of joints can be found in Sec. \ref{sec:cross_combination}
\begin{table}[!ht]
\centering
\caption{Comparison of text-condition motion generation with spacial control signal on the HumanML3D. The first section, ``Train on Pelvis Only," evaluates our model that was trained solely on the pelvis. The last section, ``Train on All Joints", is trained on all joints and assessing performance for each one. The cross-section reports performance across various combinations of joints.}
\label{tab:all_evall}
\scalebox{0.70}{
\begin{tblr}{
  colspec = {lcccccccc},
  cell{4}{2} = {r=7}{},
  cell{11}{2} = {r=3}{},
  cell{15}{2} = {r=3}{},
  cell{18}{2} = {r=3}{},
  cell{21}{2} = {r=3}{},
  cell{24}{2} = {r=3}{},
  cell{27}{2} = {r=3}{},
  cell{30}{2} = {r=2}{},
  cell{32}{2} = {r=2}{},
  hline{1-3, 11, 15, 18, 21, 24, 27, 30, 32, 34} = {-}{},
}
\textbf{Method}                  & \textbf{Joint}       & \textbf{\makecell{R-Precision \\ Top-3 $\uparrow$}} & \textbf{FID $\downarrow$}   & \textbf{\makecell{Diversity \\ $\rightarrow$}} & \textbf{\makecell{Foot Skating \\ Ratio $\downarrow$}} & \textbf{\makecell{Traj. Err. \\(50 cm) $\downarrow$}} & \textbf{\makecell{Loc. Err. \\(50 cm) $\downarrow$}} & \textbf{\makecell{Avg. Err.\\ $\downarrow$}}\\

Real                     & -      & 0.797                                 & 0.002  & 9.503     & -             & 0.0000             & 0.0000            & 0.0000    \\
\SetCell[c=9]{c,bg=gray!10} \textbf{Train on Pelvis Only} \\

MDM                     & Pelvis      & 0.602                                 & 0.698  & 9.197     & 0.1019             & 0.4022             & 0.3076            & 0.5959    \\
PriorMDM                &             & 0.583                                 & 0.475  & 9.156     & 0.0897             & 0.3457             & 0.2132            & 0.4417    \\
GMD                     &             & 0.665                                 & 0.576  & 9.206     & 0.1009             & 0.0931             & 0.0321            & 0.1439    \\
\makecell[l]{OmniControl \\ (on pelvis)}  &             & 0.687                                 & 0.218  & 9.422     & 0.0547             & 0.0387             & 0.0096            & 0.0338    \\
TLControl  & & 0.779 & 0.271  & 9.569  & - & 0.0000 & 0.0000 & 0.0108 \\
MotionLCM  & & 0.752 &  0.531 &  9.253  & -  & 0.1887 & 0.0769 &  0.1897 \\
\makecell{\textbf{MaskControl} \\ (on pelvis)}  &             & \textcolor{red}{\textbf{0.809}} & \textcolor{red}{\textbf{0.061}} & \textcolor{red}{\textbf{9.496}} & \textcolor{red}{\textbf{0.0547}} & \textcolor{red}{\textbf{0.0000}} & \textcolor{red}{\textbf{0.0000}} & \textcolor{red}{\textbf{0.0098}}   \\

\SetCell[c=9]{c,bg=gray!10} \textbf{Train on All Joints} \\
OmniControl             & Pelvis      & 0.691                                 & 0.322  & \textcolor{red}{\textbf{9.545}}     & 0.0571             & 0.0404             & 0.0085            & 0.0367    \\
TLControl  & & 0.779 &  0.271 &  9.569  & -  & 0.0000 & 0.0000 &  \textcolor{red}{\textbf{0.0108}} \\
\textbf{MaskControl}              &             & \textcolor{red}{\textbf{0.804}} & \textcolor{red}{\textbf{0.071}} & 9.453 & \textcolor{red}{\textbf{0.0546}} & \textcolor{red}{\textbf{0.0000}} & \textcolor{red}{\textbf{0.0000}} & 0.0127
   \\
OmniControl  & \makecell{Left \\ Foot}   & 0.696                                 & 0.280  & \textcolor{red}{\textbf{9.553}}     & 0.0692             & 0.0594             & 0.0094            & 0.0314    \\
TLControl  & & 0.768 & 0.368   &  9.774  & -  & 0.0000 & 0.0000 &  0.0114 \\
\textbf{MaskControl}              &             & \textcolor{red}{\textbf{0.804}} & \textcolor{red}{\textbf{0.076}} & 9.389 & \textcolor{red}{\textbf{0.0559}} & \textcolor{red}{\textbf{0.0000}} & \textcolor{red}{\textbf{0.0000}} & \textcolor{red}{\textbf{0.0072}}
   \\
OmniControl  & \makecell{Right\\Foot}  & 0.701                                 & 0.319  & \textcolor{red}{\textbf{9.481}}     & 0.0668             & 0.0666             & 0.0120            & 0.0334    \\
TLControl  & & 0.775 & 0.361 &   9.778  & -  & 0.0000 & 0.0000 &  0.0116 \\
\textbf{MaskControl}              &             & \textcolor{red}{\textbf{0.805}} & \textcolor{red}{\textbf{0.074}} & 9.400 & \textcolor{red}{\textbf{0.0549}} & \textcolor{red}{\textbf{0.0000}} & \textcolor{red}{\textbf{0.0000}} & \textcolor{red}{\textbf{0.0068}}
 \\
OmniControl             & Head        & 0.696                                 & 0.335~ & \textcolor{red}{\textbf{9.480}}     & 0.0556             & 0.0422             & 0.0079            & 0.0349    \\
TLControl  &  & 0.778 & 0.279 &  9.606  & -  & 0.0000 & 0.0000 & 0.0110  \\
\textbf{MaskControl} & & \textcolor{red}{\textbf{0.805}} & \textcolor{red}{\textbf{0.085}} & 9.415 & \textcolor{red}{\textbf{0.0538}} & \textcolor{red}{\textbf{0.0000}} & \textcolor{red}{\textbf{0.0000}} & \textcolor{red}{\textbf{0.0071}}
\\
OmniControl   & \makecell{Left\\Wrist}  & 0.680~                                & 0.304  & \textcolor{red}{\textbf{9.436}}    & 0.0562~            & 0.0801~            & 0.0134~           & 0.0529    \\
TLControl  & &  0.789  & 0.135 &  9.757  & -  & 0.0000 & 0.0000 &  0.0108 \\
\textbf{MaskControl}              &             & \textcolor{red}{\textbf{0.807}} & \textcolor{red}{\textbf{0.093}} & 9.374 & \textcolor{red}{\textbf{0.0541}} & \textcolor{red}{\textbf{0.0000}} & \textcolor{red}{\textbf{0.0000}} & \textcolor{red}{\textbf{0.0051}}
 \\
OmniControl   & \makecell{Right\\Wrist} & 0.692                             & 0.299  & \textcolor{red}{\textbf{9.519}}    & 0.0601             & 0.0813             & 0.0127            & 0.0519    \\
TLControl  & & 0.787 & 0.137  &  9.734  & -  & 0.0000 & 0.0000 & 0.0109  \\
\textbf{MaskControl}              &             & \textcolor{red}{\textbf{0.805}} & \textcolor{red}{\textbf{0.099}} & 9.340 & \textcolor{red}{\textbf{0.0539}} & \textcolor{red}{\textbf{0.0000}} & \textcolor{red}{\textbf{0.0000}} & \textcolor{red}{\textbf{0.0050}}
\\
OmniControl             & Average     & ~0.693~                               & 0.310  & \textcolor{red}{\textbf{9.502}}  & 0.0608             & 0.0617~            & 0.0107            & 0.0404    \\
\textbf{MaskControl}              &             & \textcolor{red}{\textbf{0.805}} & \textcolor{red}{\textbf{0.083}} & 9.395 & \textcolor{red}{\textbf{0.0545}} & \textcolor{red}{\textbf{0.0000}} & \textcolor{red}{\textbf{0.0000}} & \textcolor{red}{\textbf{0.0072}}

 \\
OmniControl             & Cross       & 0.672~                                & 0.624  & 9.016    & 0.0874~            & 0.2147~            & 0.0265~           & 0.0766    \\
\textbf{MaskControl}              &             & \textcolor{red}{\textbf{0.811}} & \textcolor{red}{\textbf{0.049}} & \textcolor{red}{\textbf{9.533}} & \textcolor{red}{\textbf{0.0545}} & \textcolor{red}{\textbf{0.0000}} & \textcolor{red}{\textbf{0.0000}} & \textcolor{red}{\textbf{0.0126}}
\end{tblr}}
\end{table}

\subsection{Inference speed, quality, and errors}
\label{sec:speed_quality_error}
We compare the speed of three different configurations of our model against state-of-the-art methods as shown in Table \ref{tab:speed_quality}. The first setting, \textbf{MaskControl-Fast}, uses only 100 iterations of \textit{Logits Optimization} in the last step of unmasking process. This setup achieves results comparable to OmniControl, but is over 20 times faster. It also slightly improves the Trajectory and Location Errors, while the FID score is only 25\% of OmniControl's, indicating high generation quality. The second setting, \textbf{MaskControl-Medium}, increases the \textit{Logits Optimization} to 600 iterations, which further improves accuracy. The Location Error is reduced to zero, although the FID score slightly worsens. Lastly, the \textbf{MaskControl-Accurate} model, which is the default setting used in other tables in this paper, uses 600 iterations of \textit{Logits Optimization} in the last step of unmasking process and 100 iterations of 1-9 step \textit{Logits Optimization} in the last step of unmasking process. This configuration achieves extremely high accuracy, with both the Trajectory and Location Errors reduced to zero and the Average Error below 1 cm (0.98 cm). Importantly, these settings can be adjusted during inference without retraining the model, making them suitable for both real-time and high-performance applications. Fig. \ref{tab:speed_quality} compares FID, Location Error, and speed.

\begin{table*}[!h]
\centering
\caption{Comparison of Motion Generation Performance with Speed and Quality Metrics}
\label{tab:speed_quality}
\scalebox{0.9}{
\begin{tblr}{
  colspec = {lccccccc},
  hline{1,6} = {-}{},
}
\textbf{Model} & \textbf{\makecell{Speed \\ $\downarrow$}} & \textbf{\makecell{R-Precision \\ Top-3 $\uparrow$}} & \textbf{FID $\downarrow$}   & \textbf{\makecell{Diversity \\ $\rightarrow$}} & \textbf{\makecell{Foot Skating \\ Ratio $\downarrow$}} & \textbf{\makecell{Traj. Err. \\(50 cm) $\downarrow$}} & \textbf{\makecell{Loc. Err. \\(50 cm) $\downarrow$}} & \textbf{\makecell{Avg. Err.\\ $\downarrow$}}\\
\hline
MDM                     & 10.14 s      & 0.602                                 & 0.698  & 9.197     & 0.1019             & 0.4022             & 0.3076            & 0.5959    \\
PriorMDM                &      18.11 s       & 0.583                                 & 0.475  & 9.156     & 0.0897             & 0.3457             & 0.2132            & 0.4417    \\
GMD                     & 132.49 s & 0.665                                 & 0.576  & 9.206     & 0.1009             & 0.0931             & 0.0321            & 0.1439    \\
OmniControl &    87.33 s         & 0.687                                 & 0.218  & 9.422     & 0.0547             & 0.0387             & 0.0096            & 0.0338    \\
MaskControl-Fast & 4.94 s & 0.808 & 0.059 & 9.444 & 0.0570 & 0.0200 & 0.0075 & 0.0550 \\
MaskControl-Medium & 25.23 s & 0.806 & 0.069 & 9.425 & 0.0568 & 0.0005 & 0.0000 & 0.0124 \\
MaskControl-Accurate & 71.72 s & 0.809 & 0.061 & 9.496 & 0.0547 & 0.0000 & 0.0000 & 0.0098 \\
\hline
\end{tblr}}
\end{table*}

\begin{figure}
 \centering
  \includegraphics[width=0.5\textwidth]{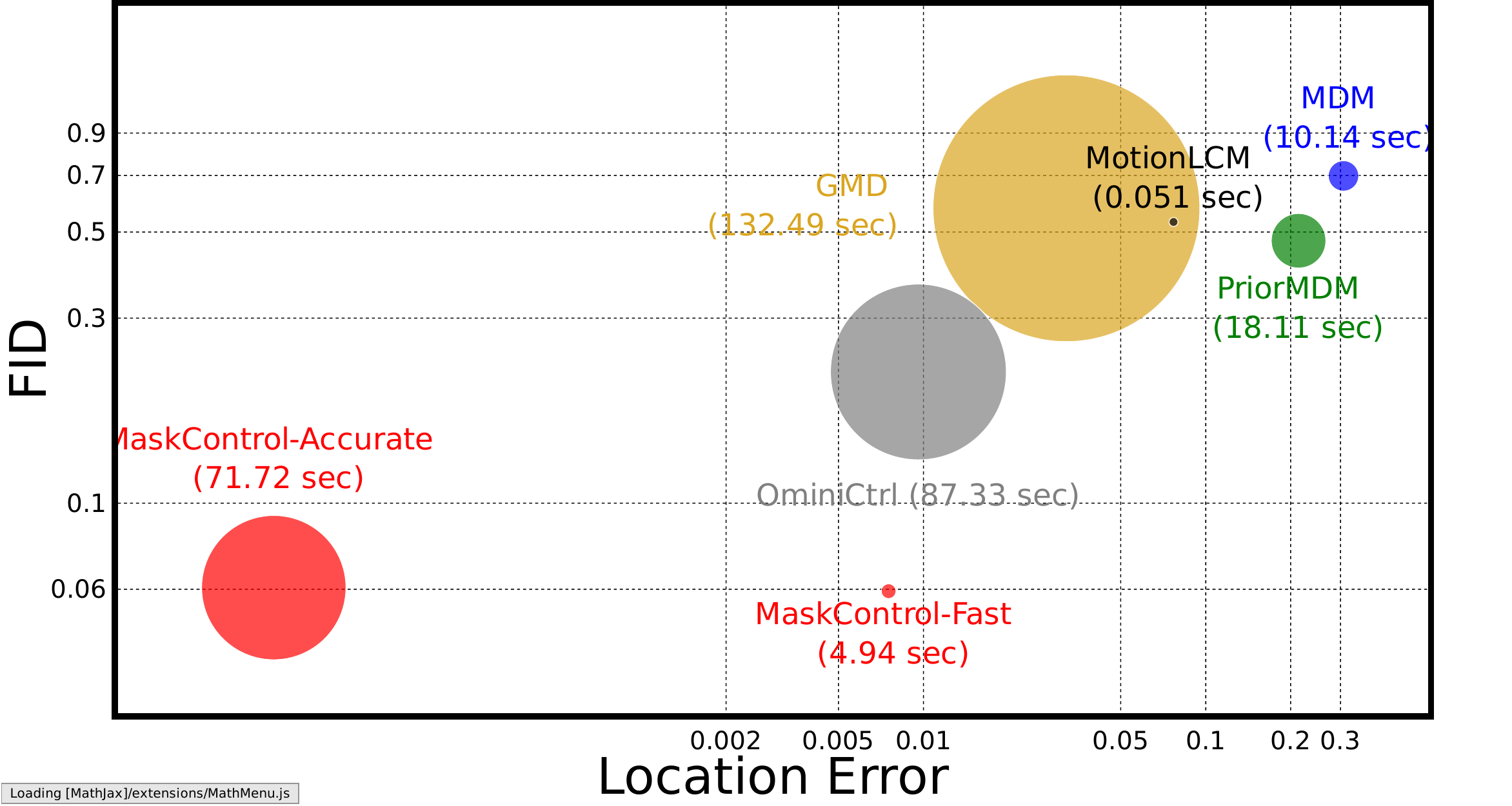}
  \caption{Comparison of FID score, spatial control error, and motion generation speed (circle size) for our accurate and fast models comparing to state-of-the-art models. The closer the point is to the origin and the smaller the circle, the better performance.}
\label{fig:speed_performance}
\end{figure}

\subsection{Speed of each component}
\label{sec:speed_component}

We report the inference time for each component in Table \ref{tab:speed_component}, with all measurements taken on an NVIDIA A100. The \textbf{Base} model, which includes only the Masked Transformer with Residual layers and Decoder (without any spatial control signal module), has an inference time of 0.35 second. The \textbf{Logits Regularizer} is highly efficient, requiring only 0.24 seconds for inference. The \textbf{Logits Optimization} takes 24.65 seconds for unmasking step 1-9 and 46.5 seconds in the last step, respectively. In total, the \textbf{MaskControl-Accurate} model has a generation time of 71.73 seconds. Note that this setting is using 100 iterations of \textbf{Logits Optimization} for 1-9 steps and 600 iterations of the last step.

\begin{table*}[!h]
\centering
\caption{Inference time of each component}
\label{tab:speed_component}
\scalebox{0.9}{
\begin{tabular}{lccccc}
\hline
 & \textbf{Base} & \textbf{Logits Regularizer} & \textbf{Logits Optimization (1-9 steps)} & \textbf{Logits Optimization (last step)} & \textbf{Full} \\
\hline
Speed in Seconds & 0.35 & 0.24 & 24.65 & 46.5 & 71.73 \\
\hline
\end{tabular}}
\end{table*}

\subsection{Quantitative result for all joints of MaskControl-Fast}
\label{sec:quantitative_result_fast}
Table \ref{tab:all_joint_fast} presents the evaluation results for MaskControl-Fast, which uses 100 iterations of \textbf{Logits Optimization}. This evaluation includes a ``cross" assessment that evaluates combinations of different joints, as detailed in Section \ref{sec:cross_combination}. The results can be compared to those of the full model (MaskControl-Accurate) and state-of-the-art models shown in Table \ref{tab:main_eval}. Additionally, "lower body" refers to the conditions involving the left foot, right foot, and pelvis, which allows for the evaluation of upper body editing tasks, as illustrated in Table \ref{tab:upperbody_editing}.

\begin{table*}[ht]
\centering
\caption{Quantitative result for all joints of MaskControl-Fast}
\label{tab:all_joint_fast}
\scalebox{0.9}{
\begin{tblr}{
  colspec = {lccccccc},
  hline{1,8,9,11} = {-}{},
  row{8} = {bg=gray!10}
}
\textbf{Joint} & \textbf{\makecell{R-Precision \\ Top-3 $\uparrow$}} & \textbf{FID $\downarrow$}   & \textbf{\makecell{Diversity \\ $\uparrow$}} & \textbf{\makecell{Foot Skating \\ Ratio $\downarrow$}} & \textbf{\makecell{Traj. Err. \\(50 cm) $\downarrow$}} & \textbf{\makecell{Loc. Err. \\(50 cm) $\downarrow$}} & \textbf{\makecell{Avg. Err.\\ $\downarrow$}}\\
\hline
 pelvis     & 0.806 & 0.067 & 9.453 & 0.0552 & 0.0446 & 0.0151 & 0.0691 \\
 left foot  & 0.806 & 0.074 & 9.450 & 0.0561 & 0.0495 & 0.0105 & 0.0484 \\
 right foot & 0.808 & 0.069 & 9.416 & 0.0566 & 0.0453 & 0.0099 & 0.0469 \\
 head       & 0.810 & 0.080 & 9.411 & 0.0555 & 0.0525 & 0.0148 & 0.0665 \\
 left wrist & 0.809 & 0.085 & 9.380 & 0.0545 & 0.0467 & 0.0108 & 0.0534 \\
 right wrist& 0.807 & 0.095 & 9.387 & 0.0549 & 0.0498 & 0.0113 & 0.0538 \\
 Average    & 0.808 & 0.079 & 9.416 & 0.0555 & 0.0481 & 0.0121 & 0.0563 \\
 cross      & 0.812 & 0.050 & 9.515 & 0.0545 & 0.0330 & 0.0101 & 0.0739 \\
 lower body & 0.807 & 0.084 & 9.396 & 0.0491 & 0.0312 & 0.0050 & 0.0633 \\
\end{tblr}}
\end{table*}

\subsection{Ablation on less number of generation step}
\label{sec:lesser_iteration}
In this section, we perform an ablation study on the number of steps used in the generation process. Following the MoMask architecture \citep{momask}, we adopt the same setting of 10 steps for generation. However, the integration of \textit{Logits Optimization} and the \textit{Logits Regularizer} enhances the quality of the generated outputs with fewer steps, as demonstrated in Table \ref{tab:less_iter}. Notably, with just 1 step, the results are already comparable to those achieved by TLControl \citep{TLcontrol}. Furthermore, after 4 steps, the evaluation metrics are on par with those obtained after 10 steps.

\begin{table}[H]
\centering
\caption{Quantitative result for different number of steps}
\label{tab:less_iter}
\scalebox{0.9}{
\begin{tblr}{
  colspec = {lccccccc},
  hline{1,8} = {-}{},
}
\textbf{\# of steps} & \textbf{\makecell{R-Precision \\ Top-3 $\uparrow$}} & \textbf{FID $\downarrow$}   & \textbf{\makecell{Diversity \\ $\rightarrow$}} & \textbf{\makecell{Foot Skating \\ Ratio $\downarrow$}} & \textbf{\makecell{Traj. Err. \\(50 cm) $\downarrow$}} & \textbf{\makecell{Loc. Err. \\(50 cm) $\downarrow$}} & \textbf{\makecell{Avg. Err.\\ $\downarrow$}}\\
\hline
 1    & 0.779 & 0.276 & 9.353 & 0.0545 & 0.0002     & 0.0000     & 0.0110  \\
 2   & 0.792 & 0.118 & 9.436 & 0.0530 & 0.0001     & 0.0000     & 0.0100 \\
 4   & 0.806 & 0.068 & 9.468 & 0.0543 & 0.0001     & 0.0000     & 0.0098 \\
 6   & 0.809 & 0.063 & 9.478 & 0.0545 & 0.0001     & 0.0000     & 0.0098 \\
 8   & 0.810 & 0.059 & 9.511 & 0.0543 & 0.0001     & 0.0000     & 0.0098 \\
10   & 0.809 & 0.061 & 9.496 & 0.0547 & 0.0000      & 0.0000    & 0.0098   \\
\end{tblr}}
\end{table}

To further investigate the influence of \textit{Logits Optimization} and the \textit{Logits Regularizer} for lesser steps, we remove these components and experiment with various numbers of steps and apply \textit{Logits Regularizer} only last step, as shown in Table \ref{tab:less_iter_noCtrlNet_noLogit}. Reducing the number of steps significantly decreases the quality of the generated outputs, resulting in an FID score of 1.196 with only 1 step. Even with 10 steps, the FID score remains at 0.190, highlighting the improvements by integrating \textit{Logits Optimization} and the \textit{Logits Regularizer}.

\begin{table}[H]
\centering
\caption{Quantitative result for different number of steps without \textit{LogitsOptimization} and \textit{Logits Regularizer}}
\label{tab:less_iter_noCtrlNet_noLogit}
\scalebox{0.9}{
\begin{tblr}{
  colspec = {lccccccc},
  hline{1,8} = {-}{},
}
\textbf{\# of steps} & \textbf{\makecell{R-Precision \\ Top-3 $\uparrow$}} & \textbf{FID $\downarrow$}   & \textbf{\makecell{Diversity \\ $\rightarrow$}} & \textbf{\makecell{Foot Skating \\ Ratio $\downarrow$}} & \textbf{\makecell{Traj. Err. \\(50 cm) $\downarrow$}} & \textbf{\makecell{Loc. Err. \\(50 cm) $\downarrow$}} & \textbf{\makecell{Avg. Err.\\ $\downarrow$}}\\
\hline
 1   & 0.716 & 1.196 & 8.831 & 0.0715 & 0.0070     & 0.0006     & 0.0271  \\
 2   & 0.758 & 0.462 & 9.182 & 0.0672 & 0.0067     & 0.0005     & 0.0276 \\
 4   & 0.782 & 0.238 & 9.236 & 0.0628 & 0.0066     & 0.0005     & 0.0281 \\
 6   & 0.787 & 0.203 & 9.276 & 0.0614 & 0.0061     & 0.0005     & 0.0282 \\
 8   & 0.787 & 0.193 & 9.272 & 0.0613 & 0.0062     & 0.0005     & 0.0283 \\
 10  & 0.786 & 0.190 & 9.294 & 0.0616 & 0.0063     & 0.0005     & 0.0283 \\
\end{tblr}}
\end{table}

\subsection{Analysis of \textit{Logits Optimization} and \textit{Logits Regularizer}}
\label{sec:analysis_of_logits}

To understand the impact of \textit{Logits Optimization} and \textit{Logits Regularizer} on the generation process, we visualize the maximum probability for each token prediction from the Masked Transformer. The model predicts 49 tokens over 10 steps. We show results both before and after applying \textit{Logits Optimization}, and with and without the \textit{Logits Regularizer}. The maximum probability can be expressed as the relative value of the logits corresponding to all codes in the codebook in the specific token position and step, as computed by the Softmax function. We visualize the output using the Softmax function instead of Gumbel-Softmax. By removing the Gumbel noise, Gumbel-Softmax reduces to a regular Softmax function:
$$
p_i = \frac{\exp(\ell_i)}{\sum_{j=1}^{k} \exp(\ell_j)}
$$
The generation is conditioned by the text prompt, ``a person walks in a circle counter-clockwise" with control over the pelvis and right hand throughout the entire trajectory. In the plot, darker blue colors represent lower probabilities (0), while yellow represents higher probabilities (1).

\textbf{Without \textit{Logits Regularizer}}

\begin{figure}[H]
 \centering
  \includegraphics[width=.5\textwidth]{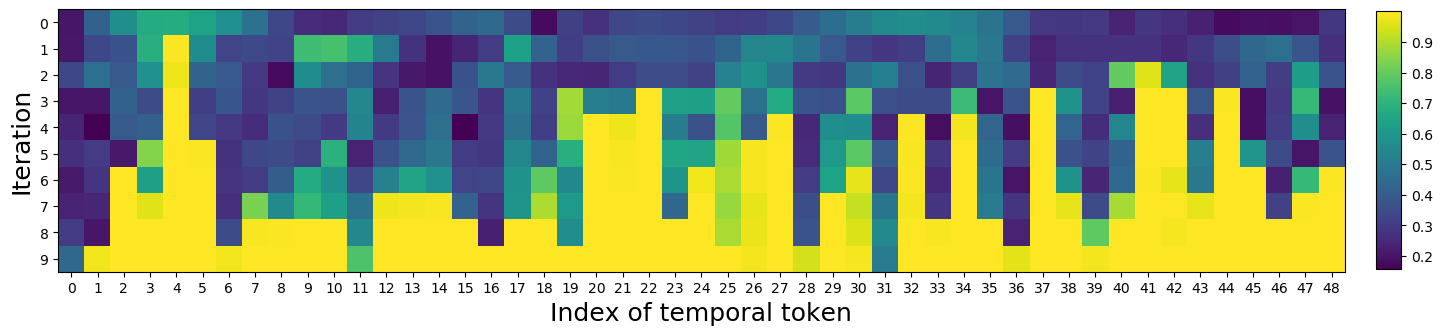}
  \caption{The maximum probability of the each token \textbf{without} \textit{Logits Regularizer} \textbf{before} \textit{Logits Optimization} of each all 49 tokens and 10 steps.}
  \label{fig:before_conf}
\end{figure}

\begin{figure}[H]
 \centering
  \includegraphics[width=.5\textwidth]{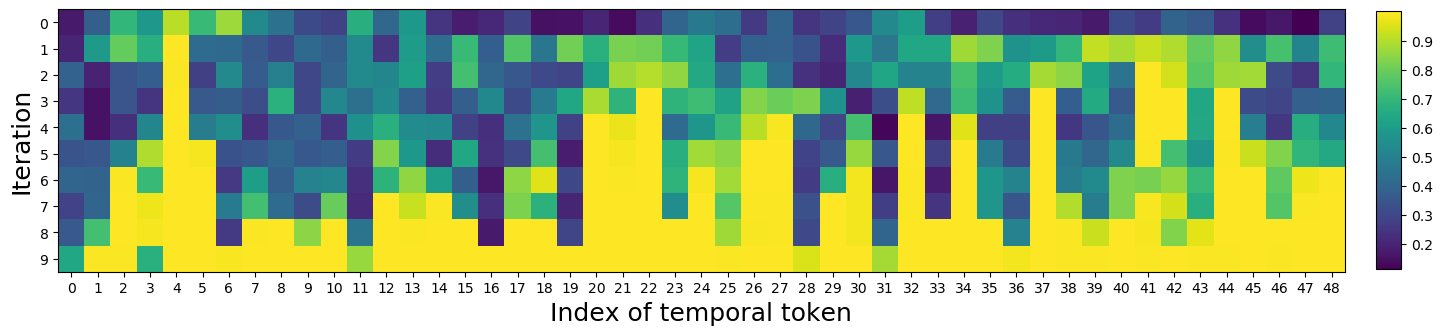}
  \caption{The maximum probability of the each token \textbf{without} \textit{Logits Regularizer} \textbf{after} \textit{Logits Optimization} of each all 49 tokens and 10 steps.}
  \label{fig:after_conf}
\end{figure}

\begin{figure}[H]
 \centering
  \includegraphics[width=.4\textwidth]{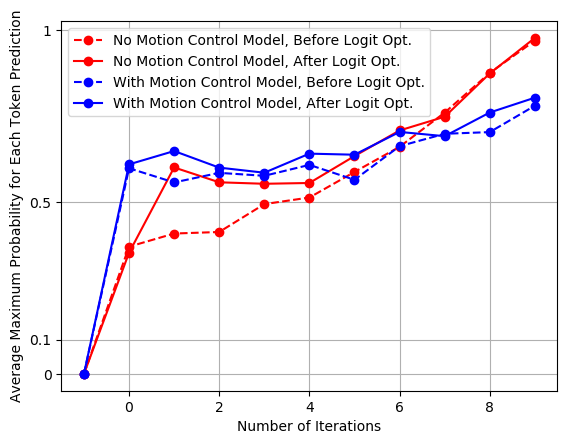}
  \caption{Average Maximum Probability for Each Token Prediction}
  \label{fig:logit_plot}
\end{figure}

In the first step (step 0), the probability is low but increases significantly in the subsequent steps. After applying \textit{Logits Optimization}, the probability improves slightly, as shown in Fig. \ref{fig:before_conf} and \ref{fig:after_conf}. Eventually, the probability saturates in the later steps (see Figure \ref{fig:logit_plot}). Since the probability of most token predictions approaches one, \textit{Logits Optimization} cannot further modify the logits, preventing any updates to the trajectory.



\textbf{With \textit{Logits Regularizer}}

\begin{figure}[H]
 \centering
  \includegraphics[width=.5\textwidth]{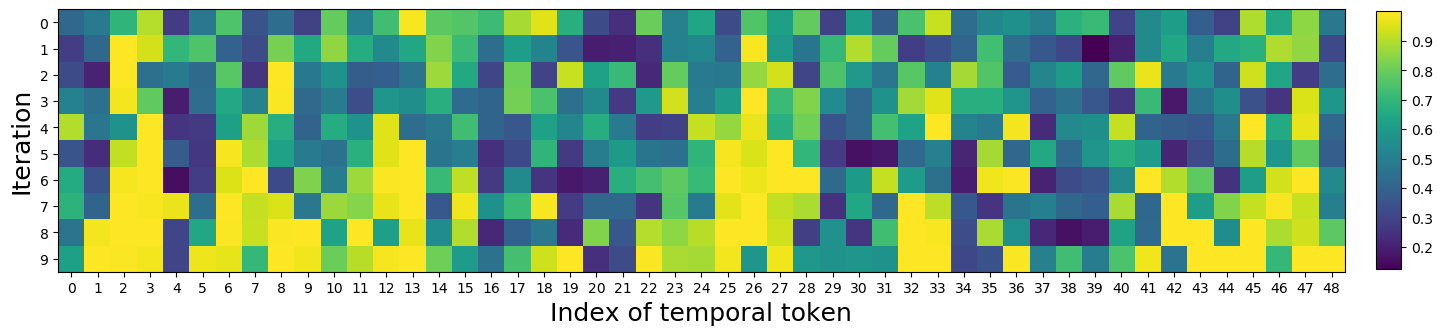}
  \caption{The maximum probability of the each token \textbf{with} \textit{Logits Regularizer} \textbf{before} \textit{Logits Optimization} of each all 49 tokens and 10 steps.}
  \label{fig:before_conf_ctrlnet}
\end{figure}

\begin{figure}[H]
 \centering
  \includegraphics[width=.5\textwidth]{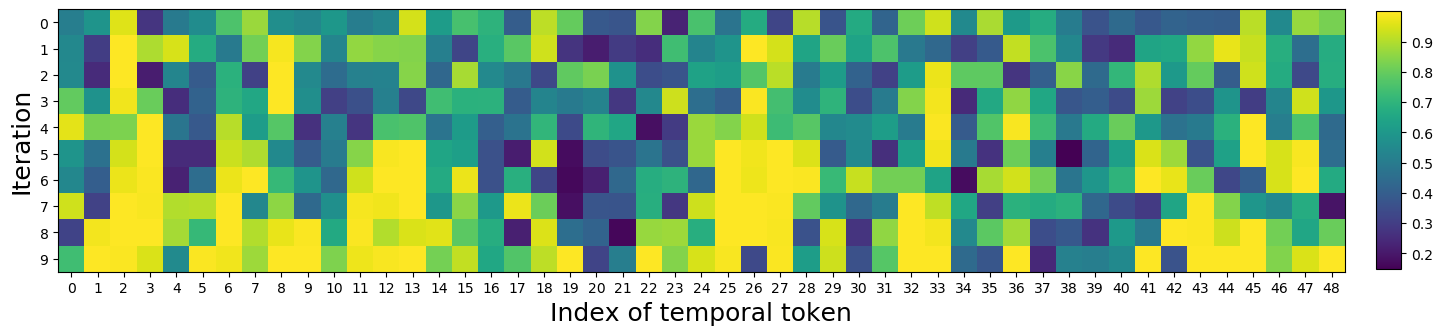}
  \caption{The maximum probability of the each token \textbf{with} \textit{Logits Regularizer} \textbf{after} \textit{Logits Optimization} of each all 49 tokens and 10 steps.}
  \label{fig:after_conf_ctrlnet}
\end{figure}

With the introduction of the \textit{Logits Regularizer}, the probability of token predictions is significantly higher in the initial step compared to the scenario without the \textit{Logits Regularizer}, as illustrated in Figures \ref{fig:before_conf_ctrlnet} and \ref{fig:after_conf_ctrlnet}. Moreover, the maximum probability does not saturate to one, indicating that there is still room to adjust the logits for trajectory editing.

This enhancement leads to improved generation quality within fewer steps, as detailed in Section \ref{sec:lesser_iteration}. Notably, just 4 steps using the \textit{Logits Regularizer} yield a quality comparable to that achieved in 10 steps without it, where the latter still exhibits suboptimal quality and high average error.

\textbf{Average of maximum probability of all tokens in each step}
To clearly illustrate the increasing probability or confidence of the model predictions across all 10 steps, as shown in Fig. \ref{fig:logit_plot}. In this figure, the blue line represents the average probability of token predictions \textcolor{blue}{\textbf{With the \textit{Logits Regularizer}}}, while the red line denotes the average probability \textcolor{red}{\textbf{Without the \textit{Logits Regularizer}}}. The solid line indicates the average probability prior to the application of \textit{Logits Optimization}. This shows that the probability increases significantly in the very first step for the \textcolor{blue}{\textbf{With the \textit{Logits Regularizer}}}.

\subsection{Ambiguity of Motion Control Signal}
\label{sec:challenges_motioncontrol}
In the image domain, pixel control signals can be directly applied, and uncontrolled regions are simply zeroed out. However, for motion control, zero-valued 3D joint coordinates are ambiguous: they may indicate that a joint is controlled with its target position at the origin in Euclidean space, or that the joint is uncontrolled. To resolve this ambiguity, we concatenate the joint control signal with the relative difference between the control signal and the generated motion, forming the final joint control guidance $s$. 


\begin{figure}[H]
 \centering
  \includegraphics[width=1\textwidth]{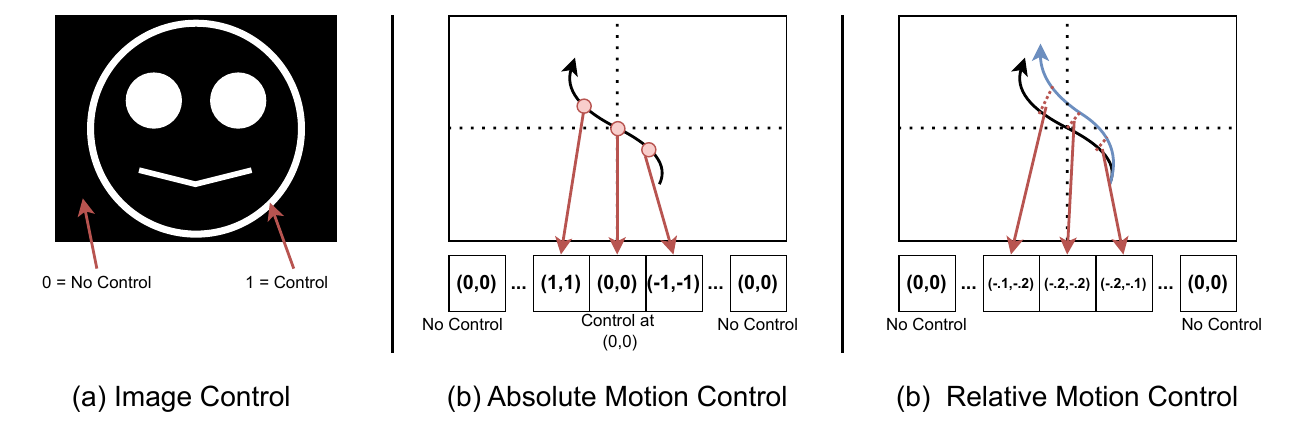}
  \caption{The difference between control signals: \textbf{(a) Image Control}: 0 means no control, 1 means control. \textbf{(b) Absolute Motion Control}: ambiguous between control signal at origin and no control. \textbf{(c) Relative Motion Control}: no ambiguity. Black curve: spatial control signal. Blue curve: decoded spatial signal from generated motion}
  \label{fig:controlnet_relative}
\end{figure}

\subsection{Body Part Timeline Control}
\label{sec:bodypart}
Generating multiple body parts based on their respective text prompts is not straightforward, as the HumanML3D dataset provides only a single prompt for each motion without specific descriptions for individual body parts. However, our model can conditionally generate outputs based on spatial signals, which allows us to manipulate and control the generation process.

\begin{figure}[H]
 \centering
  \includegraphics[width=1\textwidth]{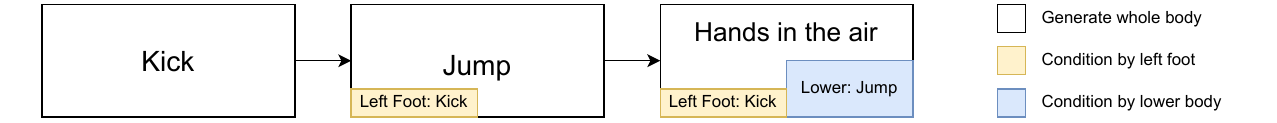}
  \caption{Process of generating body parts with multiple text inputs over a specific timeline}
  \label{fig:bodypart}
\end{figure}

To achieve this, we first generate the entire body and motion for all frames. Next, we generate a new prompt related to the next body part, using the previously generated body parts as a condition. This process can be repeated multiple times to create motion for each body part based on its corresponding text prompt, as illustrated in Fig. \ref{fig:bodypart}. 

It is important to note that this approach may lead to out-of-distribution generation since the model has not been trained on combinations of multiple body parts with their associated text prompts. However, our model handles out-of-distribution generation effectively due to the use of \textit{Logits Optimization}.

\subsection{Cross Combination}
\label{sec:cross_combination}
We follow the evaluation \textit{Cross Combination} from OmniControl \citep{OmniControl}, evaluating multiple combinations of joints as outlined in Table \ref{tab:main_eval}. A total of 63 combinations are randomly sampled during the evaluation process as follow.

\begingroup
\begin{enumerate}
    \item pelvis
    \item left foot
    \item right foot
    \item head
    \item left wrist
    \item right wrist
    \item pelvis, left foot
    \item pelvis, right foot
    \item pelvis, head
    \item pelvis, left wrist
    \item pelvis, right wrist
    \item left foot, right foot
    \item left foot, head
    \item left foot, left wrist
    \item left foot, right wrist
    \item right foot, head
    \item right foot, left wrist
    \item right foot, right wrist
    \item head, left wrist
    \item head, right wrist
    \item left wrist, right wrist
    \item pelvis, left foot, right foot
    \item pelvis, left foot, head
    \item pelvis, left foot, left wrist
    \item pelvis, left foot, right wrist
    \item pelvis, right foot, head
    \item pelvis, right foot, left wrist
    \item pelvis, right foot, right wrist
    \item pelvis, head, left wrist
    \item pelvis, head, right wrist
    \item pelvis, left wrist, right wrist
    \item left foot, right foot, head
    \item left foot, right foot, left wrist
    \item left foot, right foot, right wrist
    \item left foot, head, left wrist
    \item left foot, head, right wrist
    \item left foot, left wrist, right wrist
    \item right foot, head, left wrist
    \item right foot, head, right wrist
    \item right foot, left wrist, right wrist
    \item head, left wrist, right wrist
    \item pelvis, left foot, right foot, head
    \item pelvis, left foot, right foot, left wrist
    \item pelvis, left foot, right foot, right wrist
    \item pelvis, left foot, head, left wrist
    \item pelvis, left foot, head, right wrist
    \item pelvis, left foot, left wrist, right wrist
    \item pelvis, right foot, head, left wrist
    \item pelvis, right foot, head, right wrist
    \item pelvis, right foot, left wrist, right wrist
    \item pelvis, head, left wrist, right wrist
    \item left foot, right foot, head, left wrist
    \item left foot, right foot, head, right wrist
    \item left foot, right foot, left wrist, right wrist
    \item left foot, head, left wrist, right wrist
    \item right foot, head, left wrist, right wrist
    \item pelvis, left foot, right foot, head, left wrist
    \item pelvis, left foot, right foot, head, right wrist
    \item pelvis, left foot, right foot, left wrist, right wrist
    \item pelvis, left foot, head, left wrist, right wrist
    \item pelvis, right foot, head, left wrist, right wrist
    \item left foot, right foot, head, left wrist, right wrist
    \item pelvis, left foot, right foot, head, left wrist, right wrist
\end{enumerate}
\endgroup
\end{document}